\documentclass[reprint,aps,pre]{revtex4-1}

\usepackage{amsmath,amsfonts,amscd,amssymb,graphicx,amsthm}

\theoremstyle{definition}

\usepackage{subeqnar}
\usepackage{dcolumn}
\usepackage{booktabs}
\usepackage{overpic}
\usepackage{color}
\usepackage[caption=false]{subfig}
\usepackage{adjustbox}
\usepackage{rotating}
\usepackage{float}
\usepackage{caption}
\makeatletter
\let\newfloat\newfloat@ltx
\makeatother
\usepackage{algorithm}
\usepackage{algpseudocode}
\usepackage{comment}
\algnewcommand\algorithmicinput{\textbf{Input:}}
\algnewcommand\Input{\item[\algorithmicinput]}

\begin{document}

\title{Real-time optimal control with shallow recurrent decoder networks}

\author{Matteo Tomasetto$^\dag$, Francesco Braghin$^\dag$, J. Nathan Kutz$^{\ast}$, and Andrea Manzoni$^\ddag$}
\affiliation{$^\dag$ Department of Mechanical Engineering, Politecnico di Milano, Milano, Italy}
\affiliation{$^{\ast}$Autodesk Research, London, UK} 
\affiliation{$^{\ddag}$ MOX - Department of Mathematics, Politecnico di Milano, Milano, Italy}

\begin{abstract}
Controlling dynamical systems in real-time across multiple scenarios is critical to enabling adaptive control strategies, ensuring stability and efficiency. However, to tailor control actions in response to varying scenarios, traditional optimal control problems typically require several system simulations, which are often computationally demanding due to the high-dimensionality of the underlying spatio-temporal dynamics. In this work, we exploit SHallow REcurrent Decoder networks-based Reduced Order Modeling (SHRED-ROM) to synthesize a real-time closed-loop controller for high-dimensional and parametric dynamics, relying solely on limited state sensor readings. After training the model on a few optimal examples given by an expert demonstrator, SHRED-ROM mimics the expert behavior with effective distributed control actions in new scenarios, alleviating the curse of dimensionality. Moreover, a sensor forecaster is synthesized and used to close the loop at the latent level, thus efficiently mitigating possible sensor failures or delays. The performance of the proposed optimal control strategy is finally assessed on three challenging high-dimensional cases dealing with either parametric density control or fluid flow control.
\end{abstract}
\maketitle


\section{INTRODUCTION} \label{intro}

A wide range of phenomena in applied sciences and engineering can be described through the spatio-temporal evolution of one or more state variables which are governed by conservation laws and physical principles and expressed in terms of {\em partial differential equations} (PDEs). A model within this class that involves a (scalar-valued, for the sake of simplicity) state variable $y: \Omega \times [0,T] \to \mathbb{R}$ in the domain $\Omega$ and over the time horizon $[0,T]$ can typically be written in the form
\begin{equation}
  \dot{y}=\mathcal{N}(y,u,\mathbf{x},t;\boldsymbol{\mu}),
  \label{eq:nonlinearPDEcontrol}
\end{equation}
where $u: \Omega \times [0,T] \to \mathbb{R}$ is the (real and scalar-valued, for the sake of simplicity) control variable, $\mathcal{N}$ is the (possibly nonlinear) differential operator that models the time derivative of the state $\dot{y}$, while $\mathbf{x} \in \Omega$ and $t \in [0,T]$ are the space-time coordinates. Note that extensions to vector-valued state $\vec{y}$ and control $\vec{u}$ variables are readily possible. In addition, suitable boundary and initial conditions are required to partner with Equation~\eqref{eq:nonlinearPDEcontrol} in order to make the resulting mathematical problem well-posed. Importantly, dynamical systems in Equation~\eqref{eq:nonlinearPDEcontrol} often depend on a vector of parameters $\boldsymbol{\mu} \in \mathcal{P} \subseteq \mathbb{R}^{N_{\mu}}$, with $\mathcal{P}$ being an admissible set. Solving PDEs requires numerical methods such as, e.g., finite volume, finite element and spectral methods~\cite{kutz2013data, Quarteroni2017, courant2008methods}. In essence, through a discretization of the state $y(\mathbf{x}, t_k,\boldsymbol{\mu}) \to \mathbf{y}(t_k,\boldsymbol{\mu}) = \mathbf{y}_k^{\boldsymbol{\mu}} \in \mathbb{R}^{N_y}$ and control $u(\mathbf{x}, t_k,\boldsymbol{\mu}) \to \mathbf{u}(t_k,\boldsymbol{\mu}) = \mathbf{u}_k^{\boldsymbol{\mu}} \in \mathbb{R}^{N_u}$ over space and time -- with $t_1,\ldots,t_{N_t}$ a uniform grid over $[0,T]$ and $N_y, N_u$ being the numbers of state and control degrees of freedom -- the PDE turns into a (possibly nonlinear) system of equations to be solved over time~\cite{Quarteroni2017}, that is
\[
\mathbf{y}_{k+1}^{\boldsymbol{\mu}} = \mathbf{F}(\mathbf{y}_k^{\boldsymbol{\mu}}, \mathbf{u}_k^{\boldsymbol{\mu}}; \boldsymbol{\mu}) \quad \text{for} \; k=1,\ldots,N_t,
\]
where $\mathbf{F}:\mathbb{R}^{N_y} \times \mathbb{R}^{N_u} \times \mathcal{P} \to \mathbb{R}^{N_y}$ is the state transition map at discrete level, while $\mathbf{y}(t_1,{\boldsymbol{\mu}}) = \mathbf{y}_1^{\boldsymbol{\mu}} \in \mathbb{R}^{N_y}$ is the prescribed initial condition. Whenever a large number of degrees of freedom are required to discretize and capture the state dynamics, such as in the case of stiff phenomena or complex domains, the high-dimensionality of the resulting system entails a demanding -- or even prohibitive -- computational burden. The computational bottleneck becomes even more severe when {\em (i)} considering {\em optimal control problems} (OCPs), which require multiple PDE evaluations within an optimization loop to design optimal control actions minimizing a problem-specific loss or cost functional~\cite{kirk2004optimal, troltzsch2010optimal, manzoni2021optimal}, and {\em (ii)} parametric problems, as different independent solves are demanded in order to adapt the control actions with respect to variations in the underlying scenario. Traditional numerical methods are therefore not suitable for controlling moderate to large-scale parametric systems related to safety-critical applications with strict latency (timing) requirements, such as autonomous vehicles, robotics, plasma control, and aerospace, as delays in the control computations may imply a significant loss of performance and robustness. The goal of this work is to conceive an efficient strategy to rapidly evaluate (possibly distributed and parametric) control actions from the temporal history of limited state sensor readings, thus overcoming the computational barrier of traditional full-order solvers.

\begin{figure*}[!ht]
\captionsetup{justification=raggedright, singlelinecheck=false}
\centering
\includegraphics[width=0.9\linewidth]{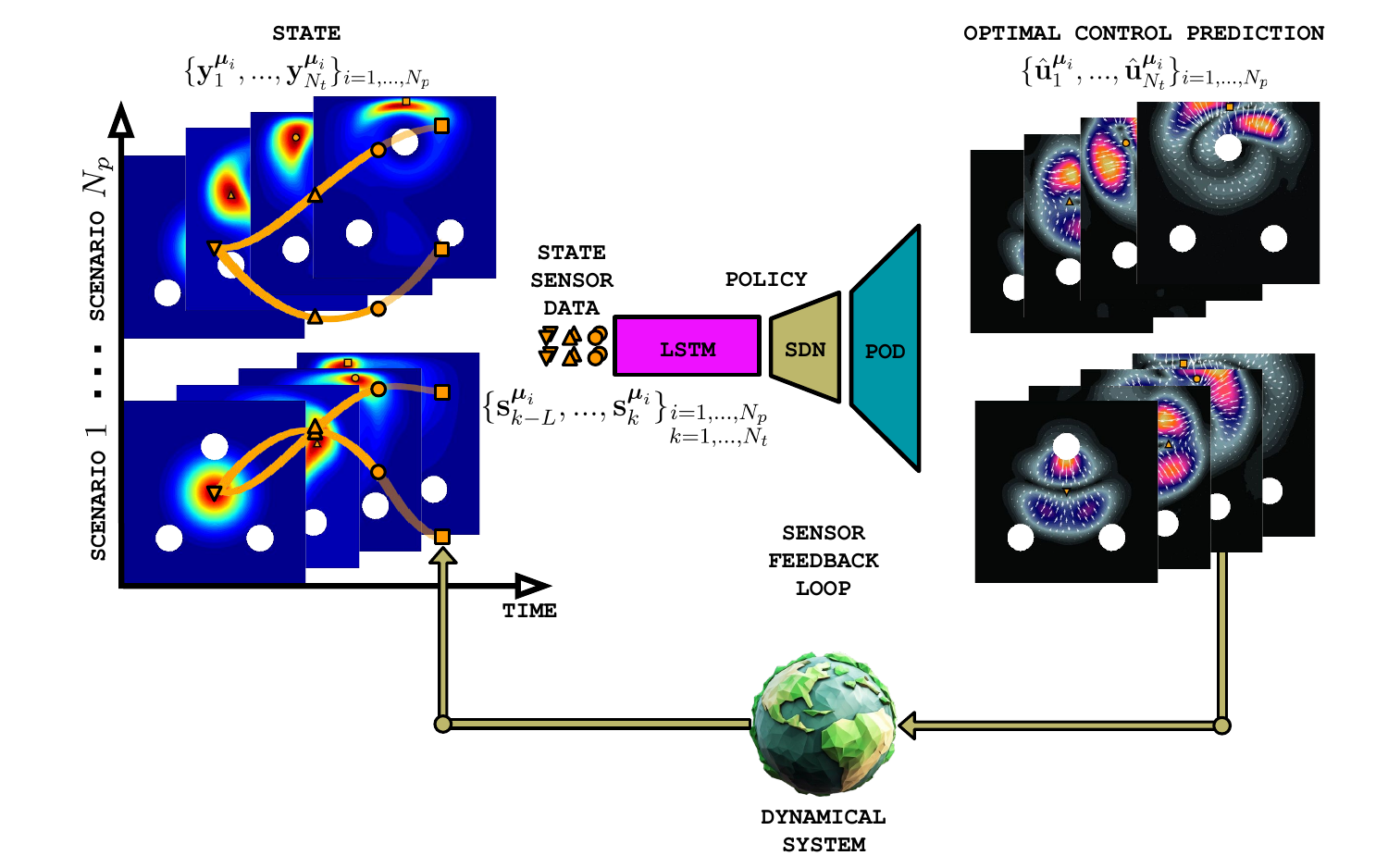}

    \caption{\textit{Graphical summary of sensor-based feedback control with SHRED-ROM}. Sparse sensor readings $\mathbf{s}_{k-L}^{\boldsymbol{\mu}_i},\ldots,\mathbf{s}_{k}^{\boldsymbol{\mu}_i}$ monitoring the state over time windows of length $L$ for $k=1,\ldots,N_t$ in multiple scenarios $i=1,\ldots,N_p$ are encoded through a long short-term memory (LSTM), while a shallow decoder network (SDN) projects the resulting latent representation in the (possibly high-dimensional) control space. The corresponding full-order control snapshots $\mathbf{u}_k^{\boldsymbol{\mu}_i}$ are reduced by proper orthogonal decomposition (POD), allowing for compressive training at the POD level. After training, in the online phase, it is possible to predict optimal control actions $\hat{\mathbf{u}}_k^{\boldsymbol{\mu}}$ for new parameters $\boldsymbol{\mu}$ unseen during training, relying solely on the past history of sparse state sensor readings. The optimal control inference may be repeated within a feedback loop in order to provide a real-time closed-loop control strategy.}
    \label{fig:shredc}
\end{figure*}
To speed up the multiple simulation requirements entailed by OCPs, it is possible to leverage accurate and fast surrogate models of the underlying dynamical system. For instance, {\em projection-based reduced order models} (ROMs) can be obtained through a (e.g., Galerkin) projection of the PDE onto a low-dimensional subspace obtained through proper orthogonal decomposition (POD)~\cite{Benner2015siamreview, kutz2013data,quarteroni2015reduced, Leibfritz2006}, thus retrieving an $r$-dimensional system to be solved over time, which is computationally tractable when $r \ll N_y$. Specifically, the {\em singular value decomposition} (SVD) of the matrix collecting the state snapshots $ \mathbf{y}_{k}^{\boldsymbol{\mu}_i} = \mathbf{y}(t_k, \boldsymbol{\mu}_i)$ for different time instants $k=1,...,N_t$ and parameter values $i = 1,...,N_p$ provides the $r$ directions of maximum variability in the data, which serve as optimal (in a least-square or statistical sense) basis for projecting the original dynamics. Alternatively, {\em physics-informed learning}~\cite{Karniadakis2021} embeds the equations describing the underlying phenomenon into the neural network training loss, thus penalizing deviations from the known physics and providing consistent state predictions. For instance, Wang et al.~\cite{Wang2021} and Hwang et al.~\cite{Hwang2022} employ physics-informed deep operator networks to learn the control-to-state map, thus enhancing optimization online. Instead of focusing on open-loop controllers, which are pre-planned strategies incapable of correcting for external disturbances or model uncertainties, the same rationale may be exploited to design closed-loop or feedback control strategies. For example, {\em model predictive control} (MPC)~\cite{Camacho2004} considers multiple open-loop physics-constrained optimization problems as the dynamics evolves over time, iteratively updating the optimal controls with respect to incoming feedback information such as, e.g., sensor data capturing the current system setting. To accelerate the multiple optimization procedures online, MPC usually employs fast-inference surrogate models --  such as, e.g., recurrent, physics-informed or echo state neural networks~\cite{Draeger1995, Aggelogiannaki2008, Bieker2019, Peitz2023, Antonelo2024, Williams2026}, POD~\cite{Ghiglieri2014, Alla2015}, or system identification techniques~\cite{Kaiser2018, Hickner2023, Korda2018, Peitz2019, Peitz2020, Klus2020} -- to quickly evaluate the state dynamics over a prediction horizon starting from the current system configuration. However, learning accurate proxies of Equation~\eqref{eq:nonlinearPDEcontrol} may be intractable when coping with distributed control actions, as significant training data would be required to adequately explore the control space. 

Rather than synthesizing surrogate models to enhance optimization, it is possible to focus on all-at-once strategies aimed at solving the {\em Karush-Kuhn-Tucker} (KKT) system of first-order optimality conditions~\cite{kirk2004optimal, manzoni2021optimal, troltzsch2010optimal}, thus avoiding optimization procedures in the online phase while allowing for large-scale actuations. For instance, projection-based ROMs~\cite{quarteroni2015reduced, Strazzullo2018,Kunisch1999,Kunisch2008,Benner2014,Amsallem2015,negri2013reduced,negri2015reduced} and physics-informed neural networks~\cite{Barry-Straume2022,Mowlavi2023,Demo2023,Yin2024,Alla2025,zhang2026pinnspdeconstrainedoptimal} may rapidly solve (online, after training) the optimality conditions, thus providing optimal state-control pairs, possibly generalizing to multiple scenarios. Similarly, feedback control design is traditionally addressed through the Hamilton-Jacobi-Bellman equation~\cite{Bardi2009, kirk2004optimal}. Despite attempts to circumvent and reduce the problem complexity in this framework -- see, e.g., \cite{Kunish2004}, \cite{Alla2016}, \cite{Schmidt2018}, \cite{Alla2020} -- numerical schemes suffer from the curse of dimensionality and become computationally intractable as the state and control dimensions increase, thus limiting their applicability in real-time control scenarios. Moreover, projection-based ROMs require complete knowledge of the underlying physics and are limited by the linearity assumption, lacking accuracy and efficiency when dealing with nonlinear and convective phenomena, while physics-informed learning typically results in oversmoothing and expensive training phases. To mitigate these limitations, {\em deep learning-based ROMs} (DL-ROMs)~\cite{Fresca2021, hesthaven2018} have been proposed as efficient nonlinear, non-intrusive and data-driven alternatives to solve high-dimensional control problems in multiple scenarios~\cite{Tomasetto2024, tomasettolatentfeedbackcontrol}. In particular, after reducing the state and control dimensionality by POD or autoencoders, the parameter-to-reduced solution map is approximated through a feed-forward neural network. While being faster and more flexible than projection-based ROMs, DL-ROMs still require full parametric knowledge to infer the state evolution, are uninformed with respect to the actual system behavior, and typically require demanding hyperparameter tuning. Conversely, since having access to sensor measurements is the standard realistic assumption in control theory, we take advantage of sensor-based reduced order models to synthesize effective control strategies that leverage this kind of information.

In this work, we employ {\em shallow recurrent decoders}~\cite{tomasetto2025shredrom, williams2024} to predict in real-time (possibly distributed) optimal control actions in multiple scenarios, relying solely on very limited feedback information -- such as, e.g., sparse sensors monitoring the state evolution -- and coping with missing or uncertain parametric knowledge. See Figure~\ref{fig:shredc} for a graphical summary of the proposed sensor-based feedback control strategy. Since continuously monitoring dynamical systems may be unfeasible due to, e.g., sensor failures or delays in communications, we also propose a computationally efficient sensor forecaster to close the loop at the latent level, thus enabling real-time optimal control strategies even in the case of missing feedback data.

The structure of the paper is as follows. Section~\ref{sec:shred-rom} recalls shallow recurrent decoders for parametric problems. Section~\ref{sec:shred-c} presents the sensor-based feedback controller, exploiting shallow recurrent decoders as control policies and recurrent neural networks as sensor forecasters to close the loop at the latent level. Finally, Section~\ref{sec:test} exploits the proposed control strategy to steer three challenging high-dimensional and parametric control problems, ranging from density control to flow control.
\vfill

\section{SHALLOW RECURRENT DECODER-BASED REDUCED ORDER MODELING}
\label{sec:shred-rom}

In this section, we review the reduced order modeling technique based on shallow recurrent decoders (SHRED-ROM) proposed in~\cite{tomasetto2025shredrom}. SHRED-ROM is a lightweight, data-driven framework aimed at reconstructing high-dimensional and parametric spatio-temporal state dynamics $\mathbf{y}_k^{\boldsymbol{\mu}} = \mathbf{y}(t_k, \boldsymbol{\mu})$ for different time instants $k=1,...,N_t$ and multiple scenarios $\boldsymbol{\mu} \in \mathcal{P}$, such as the solution of time-dependent, high-dimensional and parametric PDEs, starting from limited sensor measurements, which are often available in real-world applications. Differently from state-of-the-art sensing strategies, SHRED-ROM exploits the past history of $L$ sensor readings, i.e. 
\[
\mathbf{s}_{k-L : k}^{\boldsymbol{\mu}} = \{ 
\mathbf{s}(t_{k-L},\boldsymbol{\mu}),...,\mathbf{s}(t_k, \boldsymbol{\mu})\},
\]
where $\mathbf{s}(t_k, \boldsymbol{\mu}) \in \mathbb{R}^{N_s}$ for $k=1,\ldots,N_t$ denotes the available $N_s$ sensor values at time $t_k$ under operating condition $\boldsymbol{\mu}$, while $\mathbf{s}(t_{k-L}, \boldsymbol{\mu})=\boldsymbol{0}$ if $k \leq L$ due to pre-padding.

The possibility of reconstructing high-dimensional dynamics from sparse time-lagged sensor sequences is underpinned by the {\em Takens embedding theorem}~\cite{takens,embedology}. In essence, under mild conditions, this result guarantees that time-delayed sensor measurements are diffeomorphic to the original high-dimensional state space, i.e., the sensor sequence encodes enough dynamical information to reconstruct the underlying state up to a smooth differentiable map with differentiable inverse. The time-lagged sensor values thus represent an alternative coordinate system to the original full-order one that preserves the topology of the underlying dynamical system’s attractor. Unfortunately, the Takens embedding theorem does not provide hints on how to find the sensor-to-state map, and it can thus be approximated through data-driven approaches, such as SHRED-ROM.

SHRED-ROM is built upon the {\em shallow recurrent decoder} (SHRED) architecture, initially proposed by Williams et al.~\cite{williams2024} as a promising sparse sensing strategy in single scenarios, extending the flow reconstruction method proposed by Erichson et al.~\cite{Erichson2020} with past observations. SHRED has been then extended to encompass mobile sensors~\cite{ebers2024}, interpretable and accurate latent dynamics~\cite{gao2025sparse, yermakov2025tshredsymbolicregressionregularization}, robust predictions and uncertainty quantification~\cite{riva2024robuststateestimationpartial,gao2026uqshreduncertaintyquantificationshallow}, and data assimilation~\cite{DASHRED}. SHRED-ROM focuses on broader, more challenging, and more general problems, providing accurate approximations of high-dimensional and chaotic dynamics in multiple scenarios, such as parametric regimes never seen during training, in the direction of parametric reduced order modeling.

SHRED-ROM is a decoding-only model consisting of a sequence model $f_T$, which encodes time-lagged sensor sequences into a low-dimensional latent representation, and a decoder model $f_X$, which performs a nonlinear upscaling of the latent representation onto the original full state space, thus reconstructing the spatio-temporal quantity of interest in multiple scenarios, that is
\[
\mathbf{y}_k^{\boldsymbol{\mu}}  \approx \hat{\mathbf{y}}_k^{\boldsymbol{\mu}} = f_X(f_T(\mathbf{s}_{k-L : k}^{\boldsymbol{\mu}})) \quad  \text{for } k=1,...,N_t.
\]
In particular, we use a {\em long short-term memory network} (LSTM)~\cite{hochreiter1997long} to model the temporal dependency of sensor data, and a {\em shallow decoder network} (SDN) as latent-to-state map, even though several alternative architectures may be employed.

Whenever the state dimension $N_y$ is remarkably high, compressive training strategies may be employed to enhance computational efficiency and memory usage. Specifically, it is possible to reduce the state dimensionality through a data- or physics-driven basis expansion of the snapshots $\mathbf{y}_k^{\boldsymbol{\mu}_i}$, and train the model to its compressed representation~\cite{kutz2024shallowrecurrentdecoderreduced,tomasetto2025shredrom}. In other words, SHRED-ROM is trained to estimate only the $r \ll N_y$ basis expansion coefficients, rather than the entire high-dimensional state, enabling efficient laptop-level computing. Among different alternatives such as, e.g., Fourier modes or spherical harmonics~\cite{ye2025pyshredpythonpackageshallow, tomasetto2025shredrom}, in the following we consider POD to project the data onto a lower-dimensional subspace, that is
\[
\mathbf{y}_k^{\boldsymbol{\mu}} = \boldsymbol{\Psi}_r {\mathbf{a}_k^{\boldsymbol{\mu}}} \quad \text{for} \; k=1,\ldots,N_t,
\]
where $\boldsymbol{\Psi}_r \in \mathbb{R}^{N_y \times r}$ is the matrix collecting the first $r$ left singular vectors obtained through SVD of the state snapshot matrix, while $\mathbf{a}_k^{\boldsymbol{\mu}} \in \mathbb{R}^r$ are the corresponding basis expansion coefficients, thus yielding the reconstruction 
\begin{equation*}
    \mathbf{y}_k^{\boldsymbol{\mu}}  \approx \hat{\mathbf{y}}_k^{\boldsymbol{\mu}} = \boldsymbol{\Psi}_r\hat{\mathbf{a}}_k^{\boldsymbol{\mu}} = \boldsymbol{\Psi}_r\mathbf{f}_X(\mathbf{f}_T(\mathbf{s}_{k-L:k}^{\boldsymbol{\mu}}))  \quad  \text{for } k=1,...,N_t.
\end{equation*}
Note that, in contrast to traditional ROMs techniques, the explicit knowledge of the scenario parameters $\boldsymbol{\mu}$ is not required, as it is directly embedded in the temporal history of sensor values. Moreover, beyond reconstructing state data from its own measurements, it is possible to reconstruct one quantity from sensors monitoring a coupled field~\cite{kutz2024shallowrecurrentdecoderreduced,tomasetto2025shredrom, williams2024}. Taking advantage of this property, here we propose a strategy to synthesize parametric closed-loop controllers, enabling real-time distributed feedback control strategies across multiple scenarios in the low-data limit. 

\section{SENSOR-BASED FEEDBACK CONTROL WITH SHRED-ROM}
\label{sec:shred-c}

\begin{figure*}[!ht]
\captionsetup{justification=raggedright, singlelinecheck=false}
    \centering
    \includegraphics[width=0.9\linewidth]{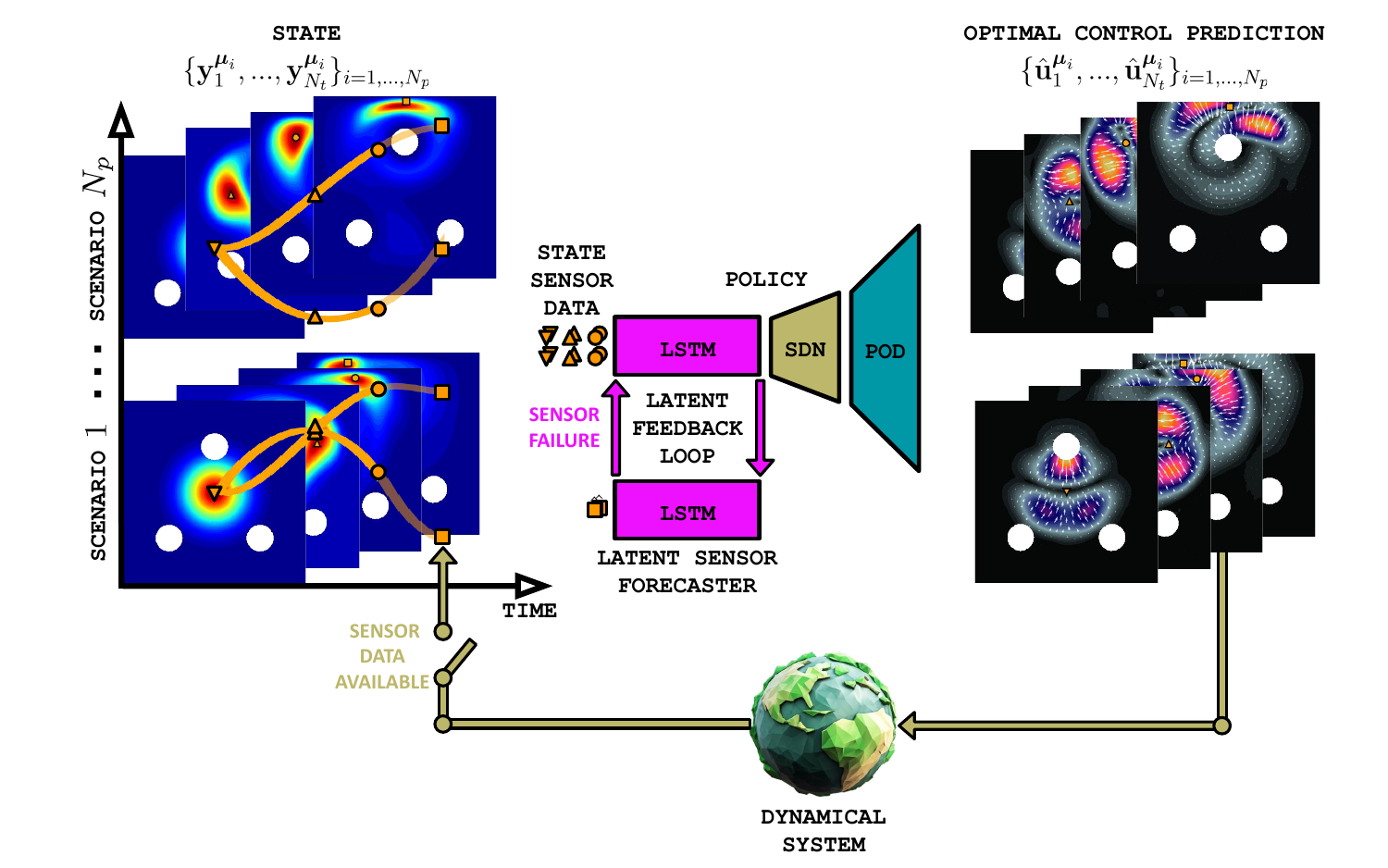}
    \caption{\textit{Graphical summary of sensor-based feedback control with latent feedback loop}. Optimal distributed control actions in multiple scenarios $\hat{\mathbf{u}}_k^{\boldsymbol{\mu}}$ are predicted in real-time relying solely on sparse state sensor readings $\mathbf{s}_{k}^{\boldsymbol{\mu}}$. The optimal control inference may be repeated within a feedback loop in order to provide a real-time closed-loop control strategy. Whenever state sensor measurements are not available online, the latent sensor forecaster is exploited to predict the sensor evolution at the latent level, allowing for a continuous control of the system.}
    \label{fig:shredclfl}
\end{figure*}
Parametric optimal control problems can be formulated as (discrete, for the sake of simplicity) PDE-constrained optimizations in the form
\begin{align*}
    &J( \mathbf{u}_k^{\boldsymbol{\mu}},  \boldsymbol{\mu}) = \sum_{k=1}^{N_t}L(\mathbf{y}_k^{\boldsymbol{\mu}},\mathbf{u}_k^{\boldsymbol{\mu}}, \boldsymbol{\mu}) + \phi(\mathbf{y}_{N_t+1}^{\boldsymbol{\mu}}, \boldsymbol{\mu}) \to \min  \\
     &\text{ s.t. } \mathbf{y}^{\boldsymbol{\mu}}_{k+1} = \mathbf{F}(\mathbf{y}^{\boldsymbol{\mu}}_k, \mathbf{u}^{\boldsymbol{\mu}}_k; \boldsymbol{\mu}) \quad \text{for} \; k=1,\ldots,N_t,
\end{align*}
where $J(\mathbf{u}_k^{\boldsymbol{\mu}},  \boldsymbol{\mu}) = J(\mathbf{u}_1^{\boldsymbol{\mu}}, \ldots, \mathbf{u}_{N_t}^{\boldsymbol{\mu}},  \boldsymbol{\mu}) \in \mathbb{R}$ is the scalar loss function to minimize by optimally designing the control variables $\mathbf{u}_k^{\boldsymbol{\mu}} = \mathbf{u}(t_k,\boldsymbol{\mu})$ for $k=1,\ldots,N_t$. Moreover, $L(\mathbf{y}_k^{\boldsymbol{\mu}}, \mathbf{u}_k^{\boldsymbol{\mu}}, \boldsymbol{\mu}) \in \mathbb{R}$ is the running cost for $k=1,\ldots,N_t$, while $\phi(\mathbf{y}_{N_t+1}^{\boldsymbol{\mu}}, \boldsymbol{\mu}) \in \mathbb{R}$ is the final payoff, where the state variables implicitly depend on the control actions through the system dynamics, i.e. $\mathbf{y}_{k+1}^{\boldsymbol{\mu}} = \mathbf{y}_{k+1}^{\boldsymbol{\mu}}(\mathbf{u}_{k}^{\boldsymbol{\mu}},\ldots, \mathbf{u}_{1}^{\boldsymbol{\mu}},\boldsymbol{\mu})$ for $k=1,\ldots,N_t$. The optimal actions over time can be computed through the Lagrangian multiplier approach or the adjoint state method~\cite{kirk2004optimal, manzoni2021optimal,troltzsch2010optimal}, yielding the KKT system of optimality conditions with state equation, adjoint equation and optimality condition, that is
\begin{equation}
\begin{cases}
 \mathbf{y}_{k+1}^{\boldsymbol{\mu}} = \mathbf{F}(\mathbf{y}_k^{\boldsymbol{\mu}}, \mathbf{u}_k^{\boldsymbol{\mu}}, \boldsymbol{\mu}) \quad & \text{for} \; k = 1,\ldots,N_t
 \vspace{0.2cm}
 \\
\boldsymbol{\lambda}_k^{\boldsymbol{\mu}} = \dfrac{\partial F}{\partial \mathbf{y}_k}^{\top} \boldsymbol{\lambda}_{k+1}^{\boldsymbol{\mu}} + \dfrac{\partial L}{\partial \mathbf{y}_k}^{\top} \quad & \text{for} \; k = N_t,\ldots,1
 \vspace{0.2cm}
 \\
\dfrac{\partial L}{\partial \mathbf{u}_k} + {\boldsymbol{\lambda}_{k+1}^{\boldsymbol{\mu}}}^{\top} \dfrac{\partial F}{\partial \mathbf{u}_k} = \boldsymbol{0} \quad & \text{for} \; k = 1,\ldots,N_t,
\end{cases}
\label{eq:kkt}
\end{equation}
where $\boldsymbol{\lambda}_k^{\boldsymbol{\mu}} \in \mathbb{R}^{N_y}$ is the discrete adjoint variable satisfying the backward-in-time adjoint equation with terminal condition $\boldsymbol{\lambda}_{N_t + 1}^{\boldsymbol{\mu}} = \frac{\partial \phi}{\partial \mathbf{y}_{N_t + 1}^{\boldsymbol{\mu}}}$. The KKT system returns, once solved, the optimal open-loop control actions over time for a fixed configuration. Different independent solves of Equation~\eqref{eq:kkt} would be required to compute the control actions for different scenarios  $\boldsymbol{\mu}$ or, after resetting time, to update them with respect to new incoming feedback information $\mathbf{y}_1^{\boldsymbol{\mu}}$. Parametric ROMs are therefore required to rapidly retrieve control actions in different settings, without degrading overall optimality, stability and system performances, while preventing damages, accidents, unsafe operations and failures. 

The coupling between state and control variables in Equation~\eqref{eq:kkt} allows us to take advantage of SHRED-ROM to build a controller providing (possibly distributed) optimal control actions in multiple scenarios starting from the temporal history of $N_s$ state sensor measurements, that is
\[
\mathbf{u}_k^{\boldsymbol{\mu}} \approx \hat{\mathbf{u}}_k^{\boldsymbol{\mu}} = f_X(f_T(\mathbf{s}_{k-L : k}^{\boldsymbol{\mu}})) \quad \text{for } k = 1,...,N_t,
\]
where $\mathbf{s}_k^{\boldsymbol{\mu}} = \{ y(\mathbf{x}_s, t_k) \}_{s=1}^{N_s} \in \mathbb{R}^{N_s}$ are state readings at some locations $\mathbf{x}_1,\ldots,\mathbf{x}_{N_s} \in \Omega$ over time $k=1,\ldots,N_t$. Importantly, pre-padding ($\mathbf{s}_k^{\boldsymbol{\mu}} = \boldsymbol{0}$ for $k \leq L$) is crucial to infer optimal control actions over the whole time horizon $[0, T]$, thus allowing for instantaneous controls with no burn-in periods. Thanks to SHRED-ROM sensor efficiency, as well as the independence on parameter values, limited state sensor readings are all you need to control high-dimensional and parametric systems in real-time. Note that, along with the optimal control prediction, the corresponding high-dimensional controlled state and the scenario parameter values may be taken into account as additional outputs of the surrogate model, as shown in Section~\ref{sec:test}, that is 
\[
\begin{bmatrix}
    \mathbf{u}_k^{\boldsymbol{\mu}} \vspace{0.1cm}\\
    \mathbf{y}_k^{\boldsymbol{\mu}}  \vspace{0.1cm}\\
    \boldsymbol{\mu}
\end{bmatrix} \approx \begin{bmatrix}
    \hat{\mathbf{u}}_k^{\boldsymbol{\mu}}  \vspace{0.1cm}\\
    \hat{\mathbf{y}}_k^{\boldsymbol{\mu}}  \vspace{0.1cm}\\
    \hat{\boldsymbol{\mu}}
\end{bmatrix} = f_X(f_T(\mathbf{s}_{k-L : k}^{\boldsymbol{\mu}})) \quad \text{for } k = 1,...,N_t.
\]

To train SHRED-ROM in a supervised manner, we consider a few optimal examples given by an expert demonstrator, as typically considered in imitation learning or behavioral cloning~\cite{hussein2018}, along with the corresponding state sensor data
\[
\mathbf{u}_k^{\boldsymbol{\mu}_i} = \mathbf{u}(t_k, \boldsymbol{\mu}_i) \in \mathbb{R}^{N_u}, \quad \mathbf{s}_k^{\boldsymbol{\mu}_i} = \{y(\mathbf{x}_s, t_k, \boldsymbol{\mu}_i)\}_{s=1}^{N_s} \in \mathbb{R}^{N_s}
\]
for different time instants $t_1,\ldots,t_{N_t}$ and scenario parameters $\boldsymbol{\mu}_1,\ldots,\boldsymbol{\mu}_{N_p} \in \mathcal{P}$. In other words, we employ SHRED-ROM to mimic the expert behavior, thus providing a real-time policy for high-dimensional and parametric dynamics in the low-data limit, while mitigating the curse of dimensionality. Note that when expanding the SHRED-ROM output with optimal state and system parameter values, the corresponding data are required during training. Note also that the need for controlled snapshots may represent a limiting factor, especially when dealing with synthetic data, and alternative strategies are required whenever such examples are not available. After splitting the available input-output pairs into training, validation, and test sets, the parameters of the sequence model $\boldsymbol{\theta}_T$ and the decoder model $\boldsymbol{\theta}_X$ are determined by minimizing the reconstruction loss
\begin{equation*}
     \mathcal{L}(\boldsymbol{\theta}_T, \boldsymbol{\theta}_X)  = \sum_{i \in \Xi_{\text{train}}} \sum_{k=1}^{N_t}  \left\lVert\mathbf{u}_k^{\boldsymbol{\mu}_i} - \mathbf{f}_X(\mathbf{f}_T(\mathbf{s}_{k-L:k}^{\boldsymbol{\mu}_i};\boldsymbol{\theta}_T);\boldsymbol{\theta}_X )\right\rVert^2
\end{equation*}
over the set of training parameter instances $\Xi_{\text{train}}$. After training, it is possible to repeatedly apply the predicted distributed control actions to the dynamical system in new scenarios unseen during training, while continuously monitoring the state through sparse sensors in order to update the steering online.

\subsection*{Latent feedback loop}

The lack of state sensor measurements online due to, e.g., sensor malfunctions or delays in communications, prevents real-time adaptation of optimal control strategies, compromising effectiveness and stability. To overcome this limitation, similarly to the loop closure at the latent level in~\cite{tomasettolatentfeedbackcontrol}, we propose a deep learning-based sensor forecaster $\varphi_z$ to predict future sensor measurements starting from their own temporal history and SHRED-ROM latent variables, that is
\begin{align*}
&\varphi_z: \underset{L \text{ times}}{\underbrace{\mathbb{R}^{N_s} \times \ldots \times \mathbb{R}^{N_s}}} \times \underset{L \text{ times}}{\underbrace{\mathbb{R}^{N_z} \times \ldots \times \mathbb{R}^{N_z}}} \to \mathbb{R}^{N_s}, \\ &{\mathbf{s}}_{k}^{\boldsymbol{\mu}} \approx \hat{\mathbf{s}}_{k}^{\boldsymbol{\mu}} = \varphi_z(\mathbf{s}_{k-L-1:k-1}^{\boldsymbol{\mu}}, \mathbf{z}_{k-L-1:k-1}^{\boldsymbol{\mu}})
\end{align*}
for $k=2,\ldots,N_t$, where $\mathbf{z}_k^{\boldsymbol{\mu}} = f_T(\mathbf{s}_{k-L:k}^{\boldsymbol{\mu}}) \in \mathbb{R}^{N_z}$ for $k=1,\ldots,N_t$ are the embeddings provided by the sequence model, while pre-padding is applied on both sensor measurements and latent control variables. Note that, for the sake of simplicity, we consider the same lag $L$ of the SHRED-ROM model, even through different values are also possible. Importantly, whenever state readings are unavailable for multiple (possibly consecutive) time steps, one can still forecast sensor values taking into account past predictions as input to $\varphi_z$, thus utilizing the latent sensor forecaster as an autoregressive model. For example, if two consecutive state sensor data are missing at time $t_{k}$ and $t_{k-1}$, we approximate $\mathbf{s}_k^{\boldsymbol{\mu}}$ with
\[
\hat{\mathbf{s}}_k^{\boldsymbol{\mu}} = \varphi_z(\mathbf{s}_{k-L-1:k-2}^{\boldsymbol{\mu}}, \hat{\mathbf{s}}_{k-1}^{\boldsymbol{\mu}}, \mathbf{z}_{k-L-1:k-1}^{\boldsymbol{\mu}}).
\]
The sensor forecaster predicts the evolution of the sensor measurements starting from their own past values, which are informative on the system configuration, as well as past latent control variables, as different control actions lead to different system evolutions. Notably, the {\em Takens embedding theorem} for non-autonomous, forced systems~\cite{Stark1999} entails that the temporal history of both sensor readings and control actions is fundamental to reconstruct the state dynamics in this context.

The latent sensor forecaster allows us to control the dynamical system even when feedback observations are compromised, thus promoting robustness, efficiency and optimality. For instance, if we experience a sensor malfunction at time $t_k$, we exploit $\varphi_z$ to approximate the sensor evolution, and we update the control strategy with
\begin{align*}
\hat{\mathbf{u}}_k^{\boldsymbol{\mu}} &= \mathbf{f}_X(\mathbf{f}_T(\mathbf{s}^{\boldsymbol{\mu}}_{k-L:k-1}, \hat{\mathbf{s}}^{\boldsymbol{\mu}}_k)) \\&= 
\mathbf{f}_X(\mathbf{f}_T(\mathbf{s}^{\boldsymbol{\mu}}_{k-L:k-1}, \varphi_z(\mathbf{s}_{k-L-1:k-1}^{\boldsymbol{\mu}}, \mathbf{z}_{k-L-1:k-1}^{\boldsymbol{\mu}}))).
\end{align*}
Importantly, whenever only a subset of sensors does not provide data in real-time, we combine the online measurements available with the $\varphi_z$ predictions of the missing one. In the following, we model the latent sensor forecaster through a LSTM with parameters $\boldsymbol{\theta}$, and we determine its weights and biases minimizing the prediction loss
\[
\mathcal{L}_z(\boldsymbol{\theta}) = \sum_{i \in \Xi_{\text{train}}} \sum_{k=1}^{N_t} \left \lVert \mathbf{s}_k^{\boldsymbol{\mu}_i} - \varphi_z(\mathbf{s}_{k-L-1:k-1}^{\boldsymbol{\mu}_i}, \mathbf{z}_{k-L-1:k-1}^{\boldsymbol{\mu}_i}) \right \rVert^2.
\]
Figure~\ref{fig:shredclfl} graphically summarizes sensor-based feedback controllers with loop closure at the latent level.

\vfill
\section{NUMERICAL RESULTS}
\label{sec:test}

This section presents the numerical results obtained when steering three high-dimensional and parametric dynamics through the proposed sensor-based feedback controllers. Since SHRED-ROM requires minimal hyperparameter tuning, we take into account the default hyperparameter values proposed in~\cite{tomasetto2025shredrom}. In particular, we split the available data into training, validation and test set with ratio $80$:$10$:$10$. Moreover, the sequence model $f_T$ shows $2$ hidden layers with $64$ neurons each, while the decoder $f_X$ is made of $2$ hidden layers having $350$ and $400$ neurons, respectively. Moreover, we select ReLU as activation function, and we prevent overfitting through dropout with rate equal to $0.1$. Regarding the neural networks training, we exploit Adam optimizer for $200$ epochs, half with learning rate equal to $0.001$ and half with learning rate equal to $0.0001$, considering a batch size equal to $64$. 

To quantitatively assess generalization capabilities of SHRED-ROM and $\varphi_z$ on new scenario parameters, we take into account the following mean relative errors
\begin{align*}
    \varepsilon(\mathbf{u}, \hat{\mathbf{u}}) &= \dfrac{1}{|\Xi_{\text{test}}|}\sum_{i \in \Xi_{\text{test}}} \sum_{k=1}^{N_t} \dfrac{\left\lVert\mathbf{u}_k^{\boldsymbol{\mu}_i} - \hat{\mathbf{u}}_k^{\boldsymbol{\mu}_i}\right\rVert}{\left\lVert\mathbf{u}_k^{\boldsymbol{\mu}_i}\right\rVert}, 
    \\
    \varepsilon(\mathbf{s}, \hat{\mathbf{s}}) &= \dfrac{1}{|\Xi_{\text{test}}|}\sum_{i \in \Xi_{\text{test}}}\sum_{k=1}^{N_t}  \dfrac{\left\lVert\mathbf{s}_k^{\boldsymbol{\mu}_i} - \hat{\mathbf{s}}_k^{\boldsymbol{\mu}_i}\right\rVert}{\left\lVert\mathbf{s}_k^{\boldsymbol{\mu}_i}\right\rVert},
\end{align*}
where $\Xi_{\text{test}}$ collects the parameter indices in the test set, while $|\Xi_\text{test}|$ denotes its cardinality. 

\subsection*{Fluidic pinball}

\begin{figure*}
    \centering
    \begin{sideways}
    \makebox[0pt][l]{\hspace{-4cm}
    \begin{minipage}{4cm}
    \centering
    {\bfseries \small UNCONTROLLED} \\ {\bfseries \small STATE} \end{minipage}}
    \end{sideways}     \subfloat[\shortstack{$1^{st}$ test scenario \\ 10 seconds}]{
    \includegraphics[width=0.21\textwidth]{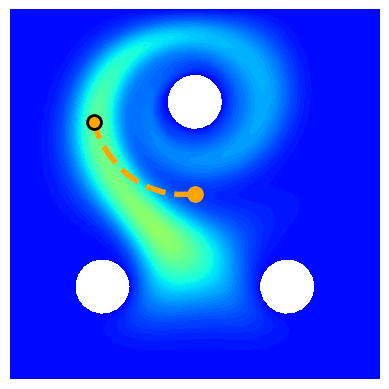}
      
    }
    \subfloat[\shortstack{$1^{st}$ test scenario \\ 30 seconds}]{
        \includegraphics[width=0.21\textwidth]{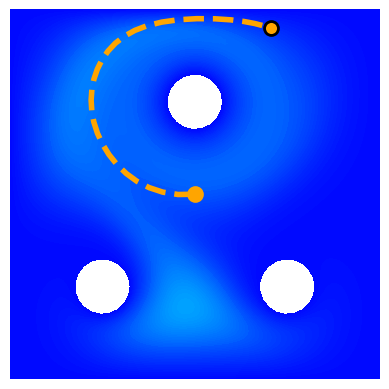}
      
    }
        \subfloat[\shortstack{$2^{nd}$ test scenario \\ 10 seconds}]{
        \includegraphics[width=0.21\textwidth]{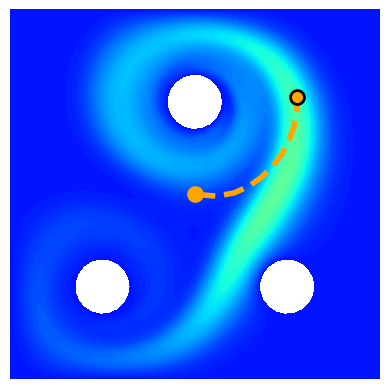}
      
    }
    \subfloat[\shortstack{$2^{nd}$ test scenario \\ 30 seconds}]{
        \includegraphics[width=0.21\textwidth]{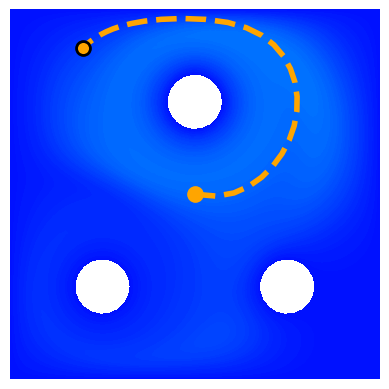}
      
    }
    \vspace{-0.3cm}

    \begin{sideways}
    \makebox[0pt][l]{\hspace{-4cm}
    \begin{minipage}{4cm}
    \centering
    {\bfseries \small TARGET} \\   {\bfseries \small STATE} \end{minipage}}
    \end{sideways}  \subfloat{
        \includegraphics[width=0.21\textwidth]{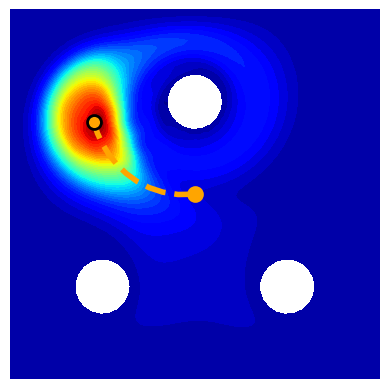}
      
    }
    \subfloat{
        \includegraphics[width=0.21\textwidth]{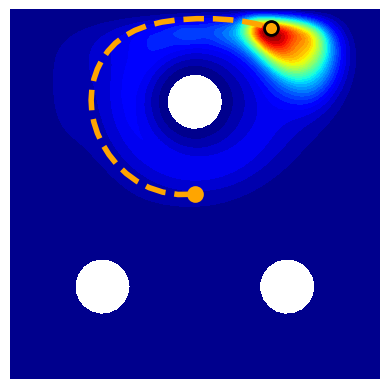}
      
    }
    \subfloat{
        \includegraphics[width=0.21\textwidth]{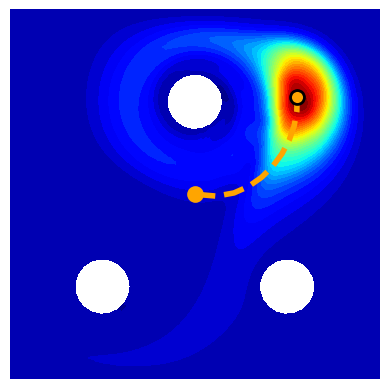}
      
    }
    \subfloat{
        \includegraphics[width=0.21\textwidth]{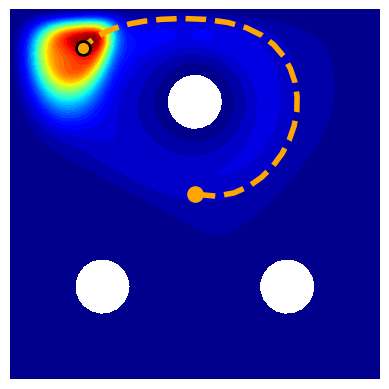}
      
    }
    \vspace{-0.3cm}

    \begin{sideways}
    \makebox[0pt][l]{\hspace{-4cm}
    \begin{minipage}{4cm}
    \centering
    {\bfseries \small CONTROLLED} \\  {\bfseries \small STATE} \end{minipage}}
    \end{sideways}  \subfloat{
        \includegraphics[width=0.21\textwidth]{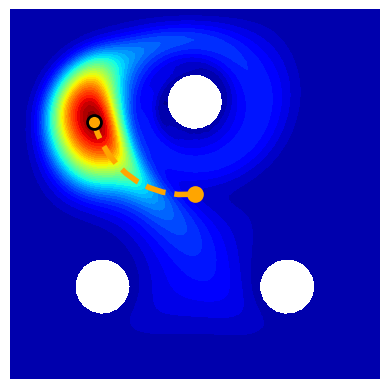}
      
    }
    \subfloat{
        \includegraphics[width=0.21\textwidth]{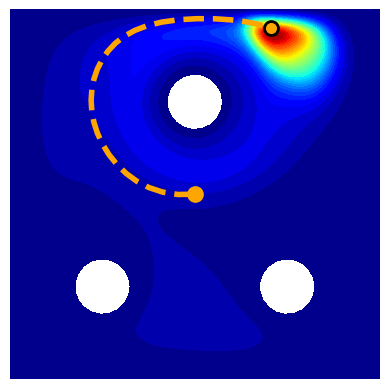}
      
    }
        \subfloat{
        \includegraphics[width=0.21\textwidth]{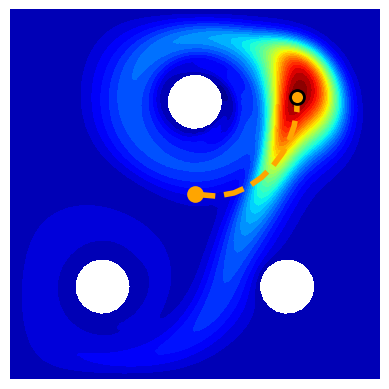}
      
    }
    \subfloat{
        \includegraphics[width=0.21\textwidth]{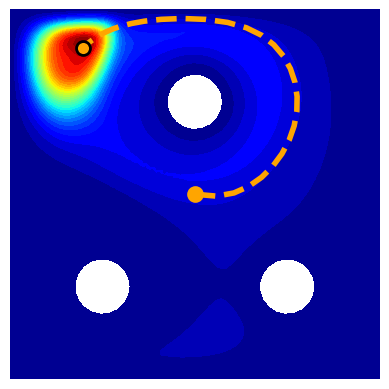}
      
    }
    \vspace{-0.3cm}

    \begin{sideways}
    \makebox[0pt][l]{\hspace{-4cm}
    \begin{minipage}{4cm}
    \centering
    {\bfseries \small TARGET} \\   {\bfseries \small CONTROL} \end{minipage}}
    \end{sideways} 
    \subfloat{
        \includegraphics[width=0.21\textwidth]{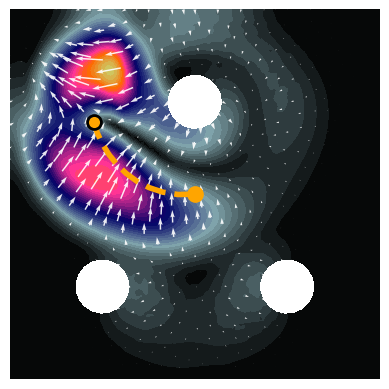}
      
    }
    \subfloat{
        \includegraphics[width=0.21\textwidth]{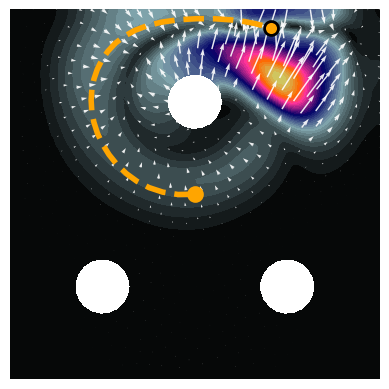}
      
    }
        \subfloat{
        \includegraphics[width=0.21\textwidth]{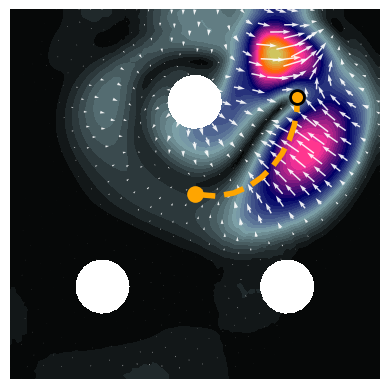}
      
    }
    \subfloat{
        \includegraphics[width=0.21\textwidth]{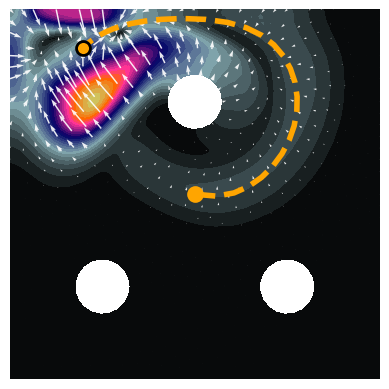}
      
    }
    \vspace{-0.3cm}

    \begin{sideways}
    \makebox[0pt][l]{\hspace{-4cm}
    \begin{minipage}{4cm}
    \centering
    {\bfseries \small SHRED-ROM} \\  {\bfseries \small CONTROL} \end{minipage}}
    \end{sideways}  \subfloat{
        \includegraphics[width=0.21\textwidth]{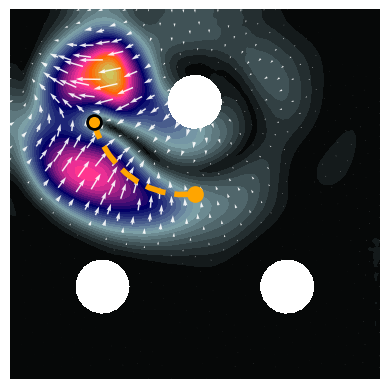}
      
    }
    \subfloat{
        \includegraphics[width=0.21\textwidth]{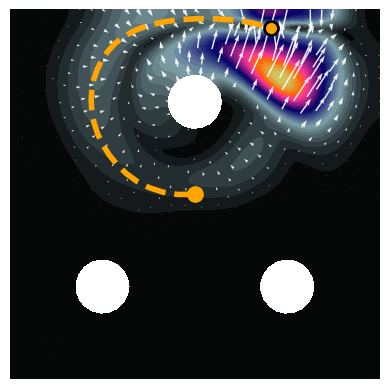}
      
    }
        \subfloat{
        \includegraphics[width=0.21\textwidth]{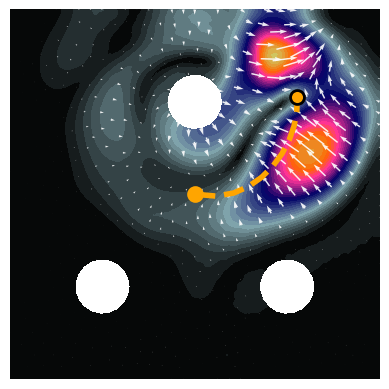}
      
    }
    \subfloat{
        \includegraphics[width=0.21\textwidth]{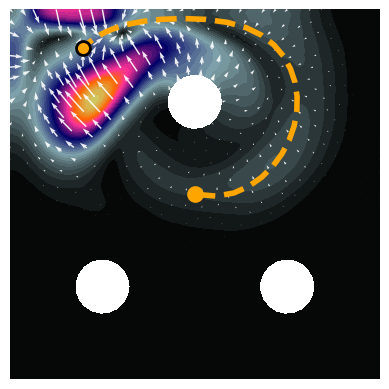}
      
    }

    \hspace{0.5cm} \subfloat{
        \includegraphics[width=0.42\textwidth]{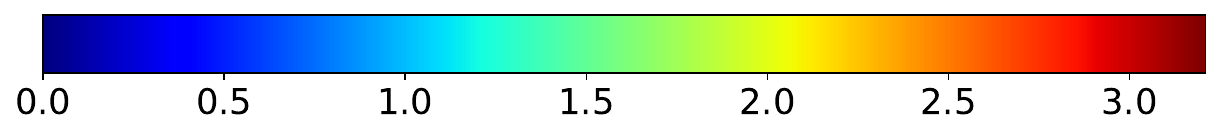}
      
    } \quad
        \subfloat{
        \includegraphics[width=0.42\textwidth]{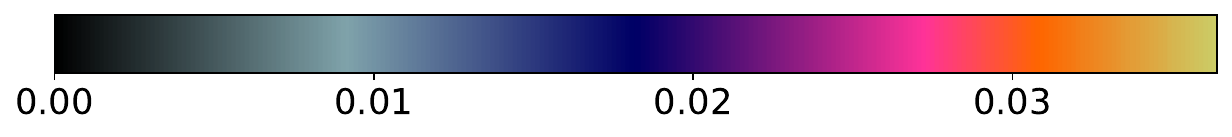}
      
    }
    \captionsetup{justification=raggedright, singlelinecheck=false}
    \caption{{\em Fluidic pinball}. Uncontrolled state (first row), optimal target state (second row), controlled state with SHRED-ROM policy (third row), target optimal control (fourth row), and SHRED-ROM control prediction (fifth row) in two different test scenarios related to $\boldsymbol{\mu} = [-0.06, 0.06, -0.70]^{\top}$ and $\boldsymbol{\mu} = [-0.74, 0.01, 0.83]^{\top}$ at $t=10$ and $t=30$ seconds. The mobile sensor trajectories are depicted in orange. The control velocity fields on $\Omega$ are depicted through vector fields, with the underlying colors corresponding to their magnitude.}
    \label{fig:pinballshredc}
\end{figure*}
\begin{figure*}
    \centering
    \subfloat[\hspace{0.5cm} $1^{st}$ test scenario]{
    \includegraphics[width=0.4\textwidth]{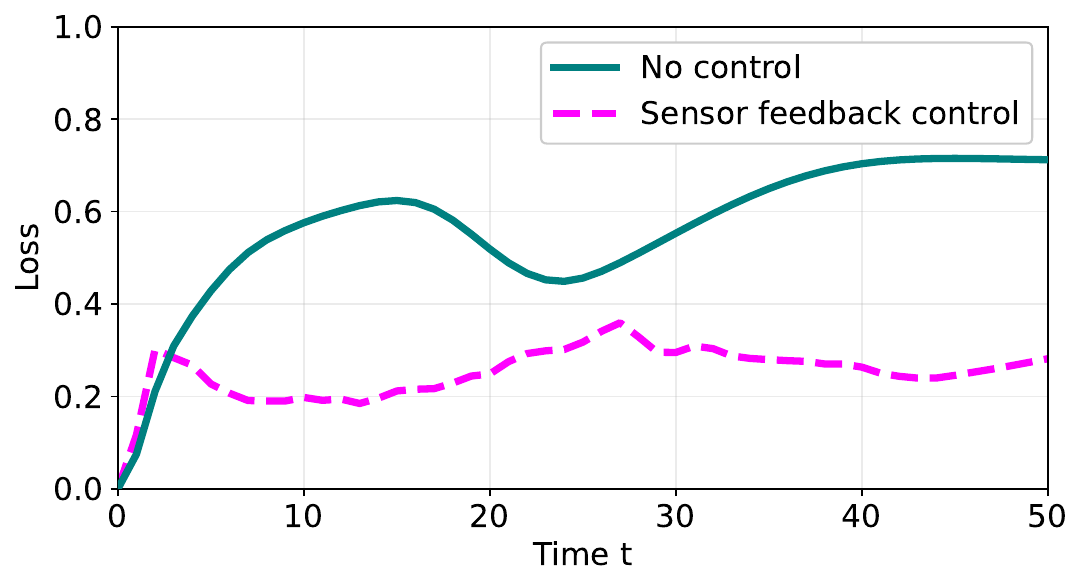}} \qquad
    \subfloat[\hspace{0.5cm} $2^{nd}$ test scenario]{
        \includegraphics[width=0.4\textwidth]{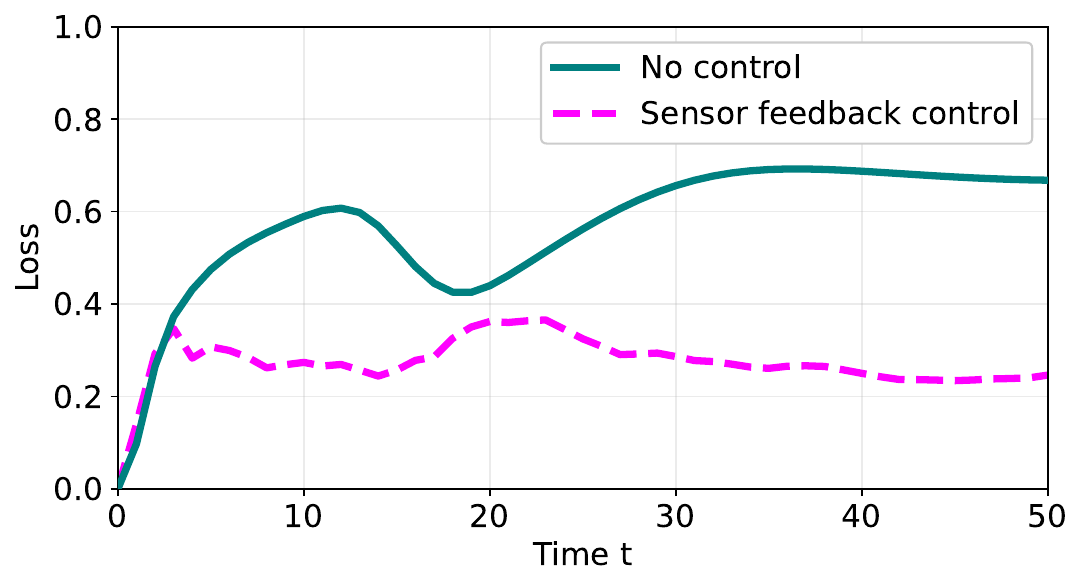}
      
    }
    \captionsetup{justification=raggedright, singlelinecheck=false}
    \caption{{\em Fluidic pinball}. Loss function over time in the uncontrolled and controlled settings with $\boldsymbol{\mu} = [-0.06, 0.06, -0.70]^{\top}$ (first panel) and $\boldsymbol{\mu} = [-0.74, 0.01, 0.83]^{\top}$ (second panel).}   
    \label{fig:pinballshredc1}
\end{figure*}
To assess the performance of the proposed controller, we first consider a high-dimensional and parametric density control problem. In particular, we take into account the fluidic pinball setting detailed in~\cite{tomasetto2025shredrom} where, starting from a Gaussian density $y_0$ centered at $(0,0)$ with variance $0.05$, the state evolves in the domain $\Omega = (-1,1)^2 \setminus \mathcal{C}_1 \cup \mathcal{C}_2 \cup \mathcal{C}_3$ -- where $ \mathcal{C}_1 = \mathcal{B}_{0.15}(-0.5,-0.5)$, $ \mathcal{C}_2 = \mathcal{B}_{0.15}(0.5,-0.5)$ and $\mathcal{C}_3 = \mathcal{B}_{0.15}(0.0,0.5)$, with $\mathcal{B}_{R}(x_1,x_2)$ denoting the cylinder centered at $(x_1,x_2)$ with radius $R$ -- according to the advection-diffusion PDE
\begin{equation}
\begin{cases}
\dot{y} + \nabla \cdot (- \eta 
\nabla y + \vec{v} y + \vec{u} y) = 0 \qquad &\mbox{in} \, \Omega \times (0,T]
\\
(- \eta \nabla y + \vec{v}y + \vec{u}y) \cdot \vec{n} = 0 \qquad &\mbox{on} \, \partial \Omega  \times (0,T]
\\
y = y_0 \qquad &\mbox{in} \, \Omega \times \{t=0\} ,
\end{cases}
\label{eq:pinballcontrol}
\end{equation}
with homogeneous Neumann boundary conditions and $\vec{n}$ being the normal unit vector. In addition to a diffusion effect with diffusivity $\eta = 0.001$, the density is transported by a fluid flow with velocity $\vec{v}:\Omega \to \mathbb{R}^2$, whose dynamics is modeled via steady Navier-Stokes equations, that is \begin{equation*}
\begin{cases}
- \nu \Delta \vec{v} + (\vec{v} \cdot \nabla) \vec{v} + \nabla p = \vec{0} &  \qquad \mbox{in} \, \Omega
\\
\nabla \cdot \vec{v} = 0& \qquad \mbox{in} \, \Omega
\\
\vec{v} = \vec{0} & \qquad \mbox{on} \, \Gamma_{\text{walls}}
\\
\vec{v}\cdot\vec{n} = 0 & \qquad \mbox{on} \, \partial \Omega
\\
\vec{v}\cdot\vec{t} = v_1 & \qquad \mbox{on} \, \partial \mathcal{C}_1
\\
\vec{v}\cdot\vec{t} = v_2 & \qquad \mbox{on} \, \partial \mathcal{C}_2
\\
\vec{v}\cdot\vec{t} = v_3 & \qquad \mbox{on} \, \partial \mathcal{C}_3,
\end{cases}
\end{equation*}
with kinematic viscosity $\nu = 1.0$, $p:\Omega \to \mathbb{R}$ the pressure variable, $\vec{t}$ the tangential unit vector, and no-slip boundary conditions on the external walls $\Gamma_{\text{walls}}$. The fluid velocity implicitly depends on three rotating cylinders in the domain, whose constant velocities are regarded as tangential Dirichlet boundary data and scenario parameters, that is $\boldsymbol{\mu} = [v_1, v_2, v_3]^\top$. 

As visible in the first row of Figure~\ref{fig:pinballshredc}, the uncontrolled setting is characterized by significant density dispersion throughout the domain, as well as collisions with the boundaries. Preventing these phenomena in multiple scenarios -- i.e. for different combinations of cylinder velocities -- would be beneficial whenever the density $y$ serves as the macroscopic description of, e.g., robotic swarms or pollutant concentration. To do so, we take into account a distributed velocity field $\vec{u}$ in Equation~\eqref{eq:pinballcontrol}, whose values have to be optimally designed in order to minimize the loss function
\begin{equation*}
\begin{aligned}
J(\vec{u}, \boldsymbol{\mu}) &= \frac{1}{2} \int_0^{T} \int_{\Omega} (y - \bar{y})^2 d\Omega dt + \int_0^{T} \int_{\partial \Omega} y^2 d\Gamma dt + 
\\
&+ \frac{\beta}{2} \int_0^{T} \int_{\Omega} ||\vec{u}||^2 d\Omega dt + \frac{\beta_g}{2}  \int_0^{T} \int_{\Omega} ||\nabla \vec{u}||^2 d\Omega dt, 
\end{aligned}
\end{equation*}
where $\beta=10^{-4}$ penalizes control effort, while $\beta_g= 10$ promotes control smoothness. To avoid density dispersion while following the fluid flow, the target state $\bar{y}(t)$ is defined as the Gaussian density
\[
\bar{y}(t) = \dfrac{10}{\pi} \exp{(- 10(x_1 - \bar{\mu}_1(t))^2 - 10(x_2 - \bar{\mu}_2(t))^2)}
\]
with constant variance equal to $0.05$ and mean position $\bar{\mu}(t) = (\bar{\mu}_1(t),\bar{\mu}_2(t))$ given by the ODE
\begin{equation*}
\begin{cases}
\dot{\bar{\boldsymbol{\mu}}}(t) = \vec{v}(\bar{\boldsymbol{\mu}}, t) 
\\
\bar{\boldsymbol{\mu}}(0) = \vec{0}.
\end{cases}
\end{equation*}
Moreover, to avoid collisions with the domain boundaries, we minimize the $L^2$ norm of the density over $\partial \Omega \times [0,T]$.

To synthesize a sensor-based feedback controller, we first discretize the forward problem with finite element method, resulting in full-order state and control dimensions equal to, respectively, $N_y = 7525$ and $N_u = 59344$, while we employ an evenly spaced time discretization with $\Delta t= 1$ second and $T=50$ seconds. We then generate $N_p = 100$ optimal examples through the full-order adjoint method~\cite{Mitusch2019} for different cylinder velocities randomly sampled in the parameter space $\mathcal{P} = [-1,1]^3$, and we split the optimal trajectories into training, validation and test sets with ratio $80$:$10$:$10$. Finally, we train the SHRED-ROM policy to predict the optimal control velocity field starting from the $L=10$ past state measurements and the coordinates of $N_s=1$ mobile sensor, which is placed in the center of the domain at $t=0$ and is passively steered by the underlying fluid flow velocity $\vec{v}$. Alternatively, sparse fixed sensors monitoring the state evolution may be considered.

To speed up training, it is possible to employ compressive training strategies where the control snapshots are reduced through $r=300 \ll N_u$ POD modes. After training the sensor-based feedback controller, it is possible to deploy the synthesized policy in test scenarios unseen during training. Figure~\ref{fig:pinballshredc} and Figure~\ref{fig:pinballshredc1} qualitatively and quantitatively assess the performance of the SHRED-ROM policy in two test scenarios, with effective minimization of density dispersion and boundary collisions compared to the uncontrolled setting, achieving performance comparable to the full-order target optimal solutions in the test set.

\subsection*{Unsteady flow control}
\label{subsec:flowcontrolshredc}

In this section, we test the proposed sensor-based feedback control strategy in the high-dimensional and parametric unsteady flow control problem introduced in~\cite{Tomasetto2024}. In particular, we take into account a fluid flow in a rectangular channel with a circular obstacle inside, that is $\Omega = (0,8) \times (0,2) \setminus \mathcal{B}_{0.2}(1,1)$, whose dynamics is described by the unsteady Navier-Stokes equations

\begin{figure*}
    \centering
    \begin{sideways}
    \makebox[0pt][l]{\hspace{-2.75cm}
    \begin{minipage}{4cm}
    \centering
    {\bfseries \small NO} \\ {\bfseries \small CONTROL} \end{minipage}}
    \end{sideways}     \subfloat[\shortstack{$1^{st}$ test scenario \\ 4 seconds}]{
    \includegraphics[width=0.3\textwidth]{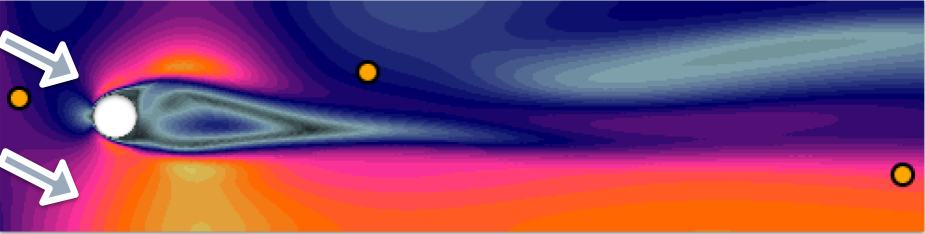}
      
    }
    \subfloat[\shortstack{$2^{nd}$ test scenario \\ 4 seconds}]{
        \includegraphics[width=0.3\textwidth]{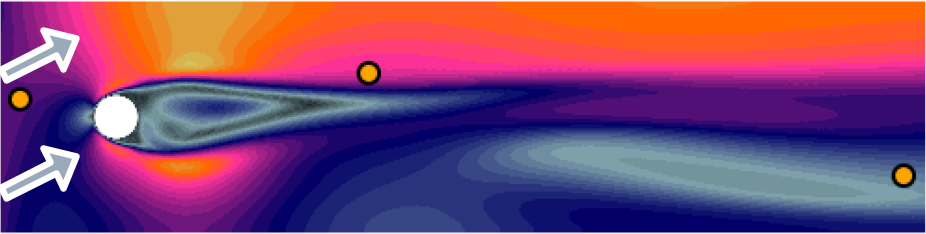}
      
    }
        \subfloat[\shortstack{$3^{rd}$ test scenario \\ 4 seconds}]{
        \includegraphics[width=0.3\textwidth]{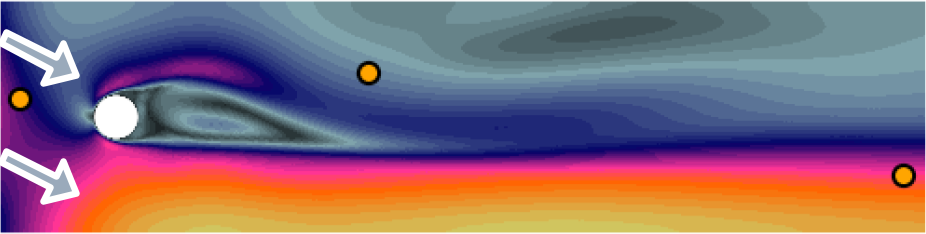}
      
    }
    \vspace{0.4cm}

    \begin{sideways}
    \makebox[0pt][l]{\hspace{-3.4cm}
    \begin{minipage}{4cm}
    \centering
    {\bfseries \small TARGET} \\   {\bfseries \small PAIR} \end{minipage}}
    \end{sideways}  \subfloat{
        \includegraphics[width=0.3\textwidth]{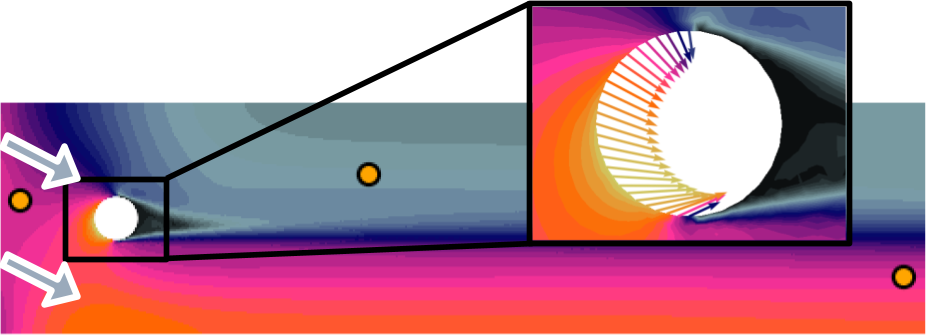}
      
    }
    \subfloat{
        \includegraphics[width=0.3\textwidth]{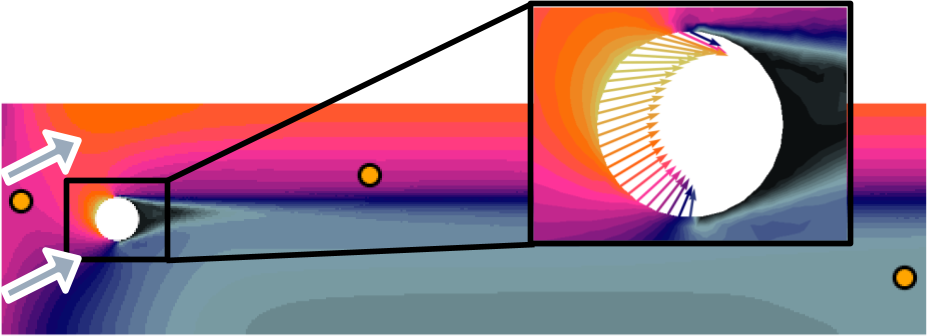}
      
    }
    \subfloat{
        \includegraphics[width=0.3\textwidth]{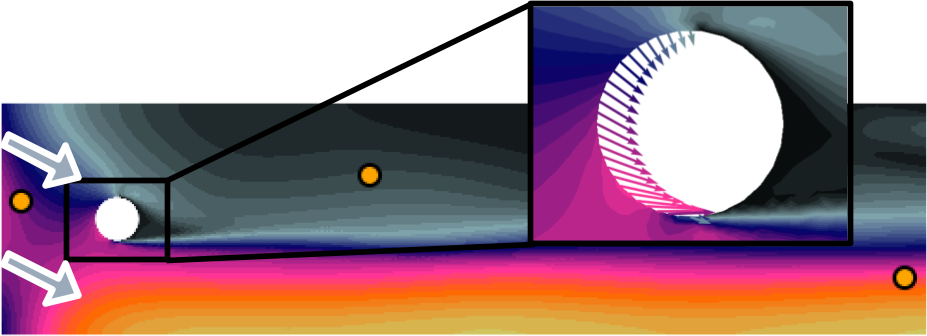}
      
    }
    \vspace{0.4cm}

    \begin{sideways}
    \makebox[0pt][l]{\hspace{-5cm}
    \begin{minipage}{4.5cm}
    \centering
    {\bfseries \small SENSOR-BASED} \\  {\bfseries \small FEEDBACK CONTROL} \end{minipage}}
    \end{sideways}  \subfloat{
        \includegraphics[width=0.3\textwidth]{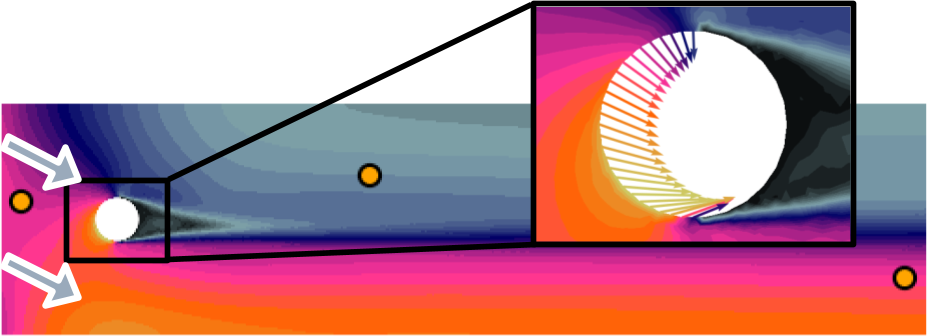}
      
    }
    \subfloat{
        \includegraphics[width=0.3\textwidth]{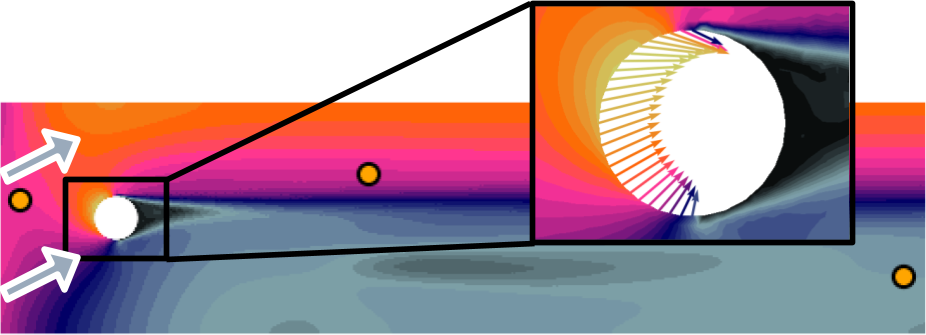}
      
    }
        \subfloat{
        \includegraphics[width=0.3\textwidth]{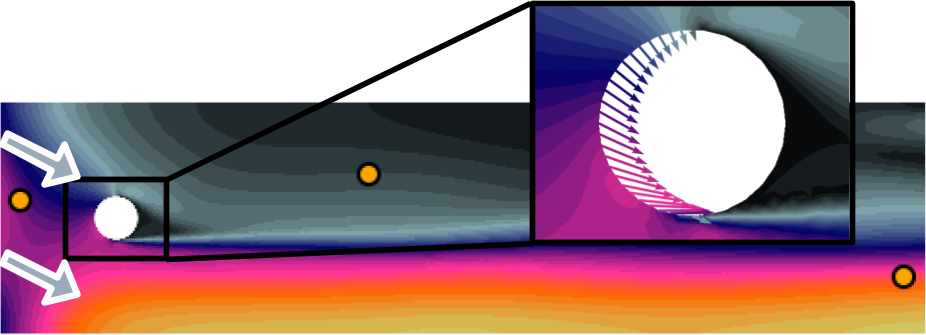}
    }
    \vspace{0.4cm}

    \begin{sideways}
    \makebox[0pt][l]{\hspace{-4.25cm}
    \begin{minipage}{4cm}
    \centering
    \vphantom{\bfseries \small SHRED-ROM} \\   \vphantom{\bfseries \small PREDICTION} \end{minipage}}
    \end{sideways} 
    \subfloat{
        \includegraphics[width=0.3\textwidth]{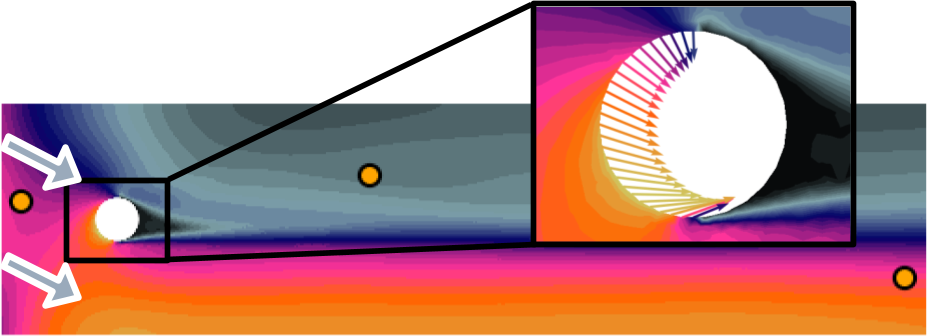}
      
    }
    \subfloat{
        \includegraphics[width=0.3\textwidth]{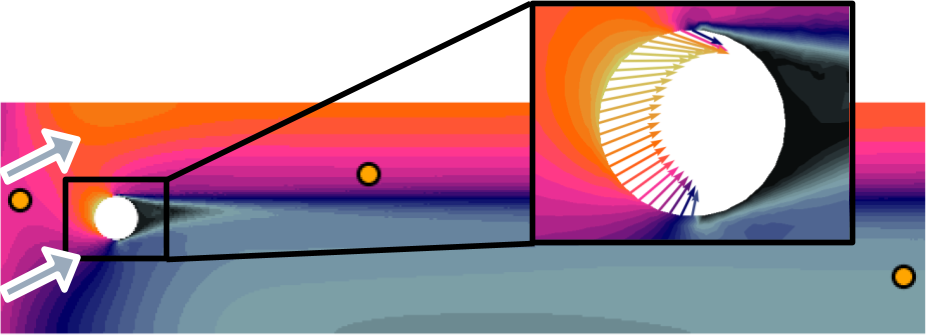}
      
    }
        \subfloat{
        \includegraphics[width=0.3\textwidth]{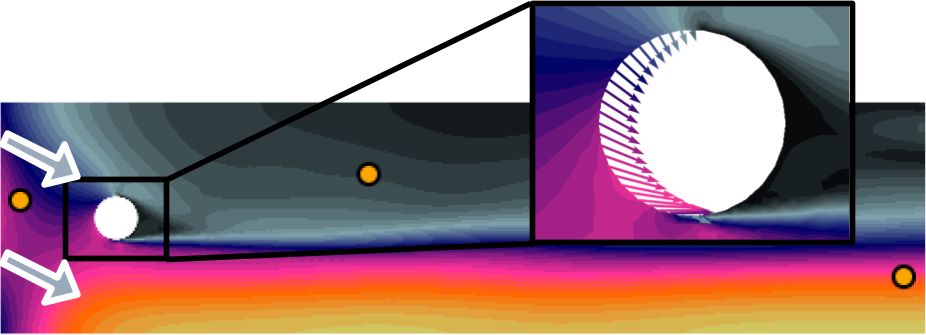}
      
    }

    \begin{sideways}
    \makebox[0pt][l]{\hspace{-4.75cm}
    \begin{minipage}{4cm}
    \centering
    {\bfseries \small LATENT} \\  {\bfseries \small FEEDBACK LOOP} \end{minipage}}
    \end{sideways}  \subfloat{
        \includegraphics[width=0.3\textwidth]{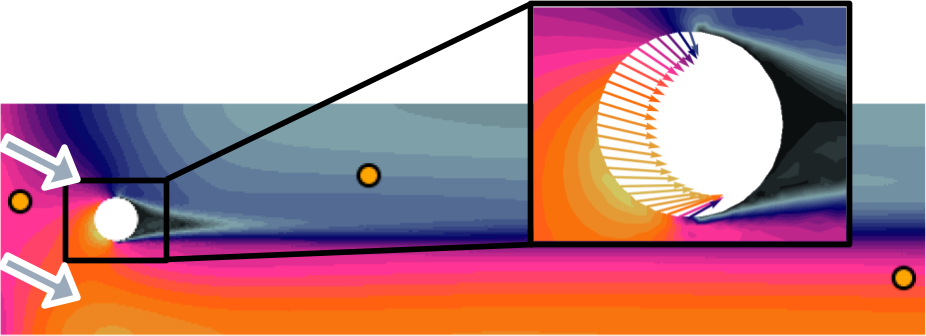}
      
    }
    \subfloat{
        \includegraphics[width=0.3\textwidth]{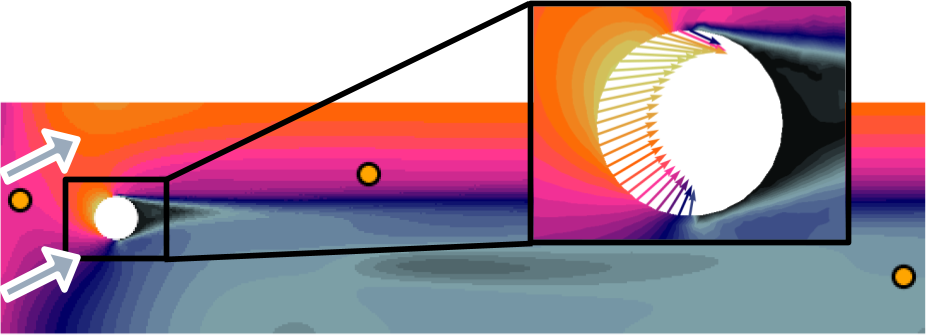}
      
    }
        \subfloat{
        \includegraphics[width=0.3\textwidth]{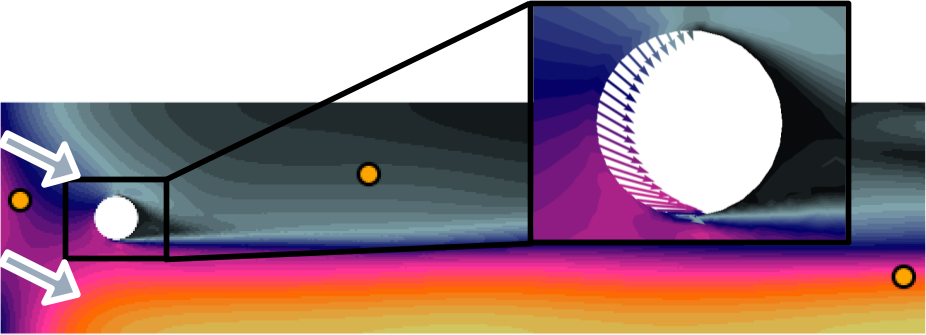}
    }
    \vspace{0.4cm}

    \begin{sideways}
    \makebox[0pt][l]{\hspace{-3cm}
    \begin{minipage}{4cm}
    \centering
    \vphantom{\bfseries \small SHRED-ROM} \\   \vphantom{\bfseries \small PREDICTION} \end{minipage}}
    \end{sideways} 
    \subfloat{
        \includegraphics[width=0.3\textwidth]{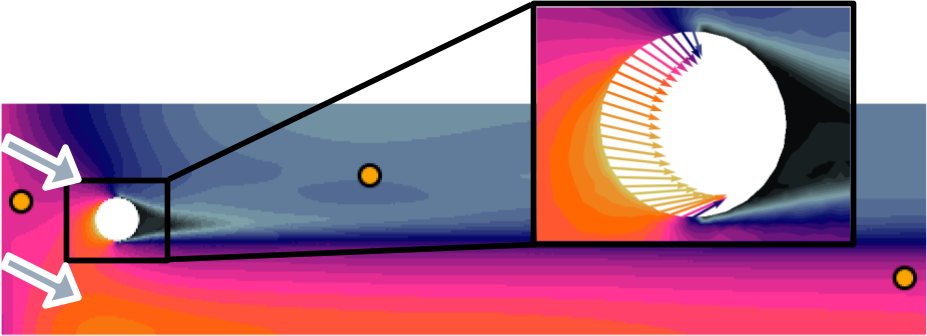}
      
    }
    \subfloat{
        \includegraphics[width=0.3\textwidth]{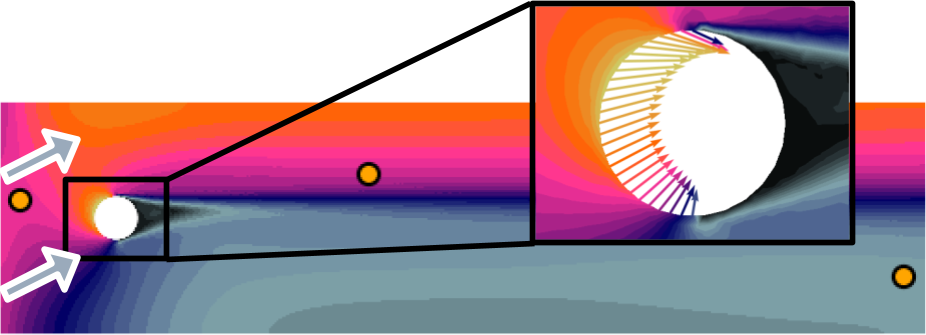}
      
    }
        \subfloat{
        \includegraphics[width=0.3\textwidth]{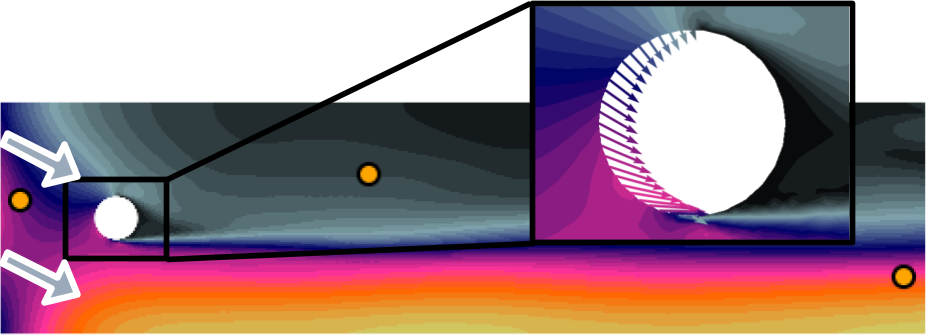}
      
    }

    \vspace{0.1cm}
    \hspace{0.55cm}
    \subfloat{
        \includegraphics[width=0.4\textwidth]{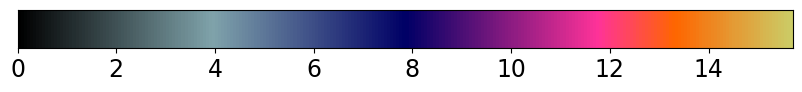}
      
    }
    \captionsetup{justification=raggedright, singlelinecheck=false}
    \caption{{\em Unsteady flow control}. Uncontrolled velocity (first row), optimal target pair (second row), controlled velocity and optimal pair prediction with sensor-based feedback control (third and fourth row), controlled velocity and optimal pair prediction with latent feedback loop (fifth and sixth row) in two different test scenarios related to $\mu = -0.39, 0.39,-0.76$ at $t=4$ seconds. The fixed sensors are depicted with orange dots. The velocity on $\Omega$ is depicted through a scalar field with colors corresponding to its norm, while the boundary control is represented through a vector field.}
    \label{fig:flowshredc}
\end{figure*}

\begin{figure*}
    \centering
    \subfloat[\hspace{0.5cm} $1^{st}$ test scenario]{
    \begin{overpic}[width=0.32\textwidth]{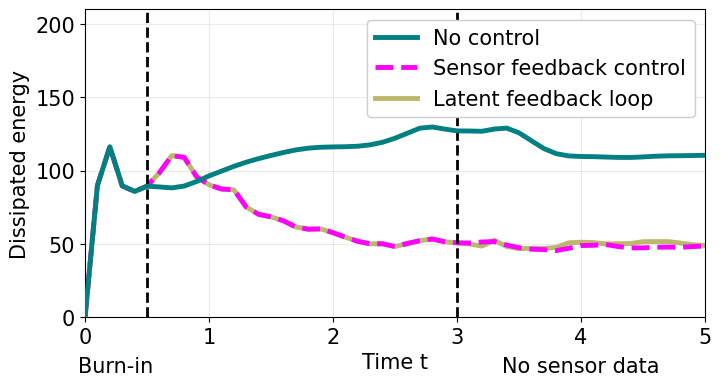}\put(-10,50){\bfseries (a)}\end{overpic}}
    \subfloat[\hspace{0.5cm} $2^{nd}$ test scenario]{
        \includegraphics[width=0.32\textwidth]{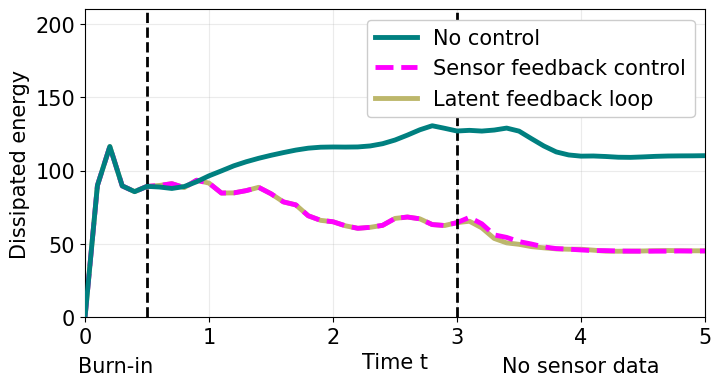}
      
    }
        \subfloat[\hspace{0.5cm} $3^{rd}$ test scenario]{
        \includegraphics[width=0.32\textwidth]{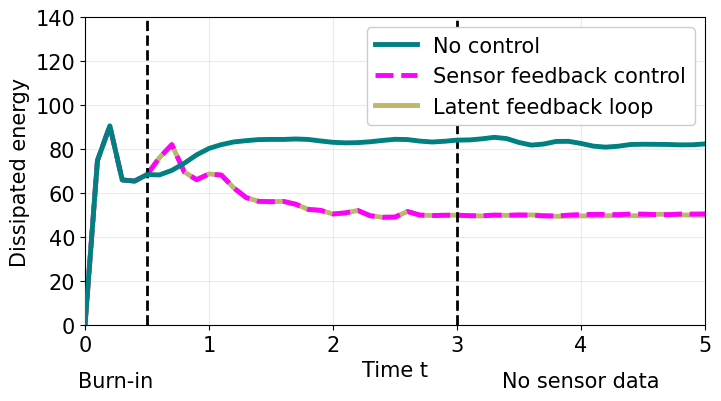}
      
    }

    \subfloat{
    \begin{overpic}[width=0.32\textwidth]{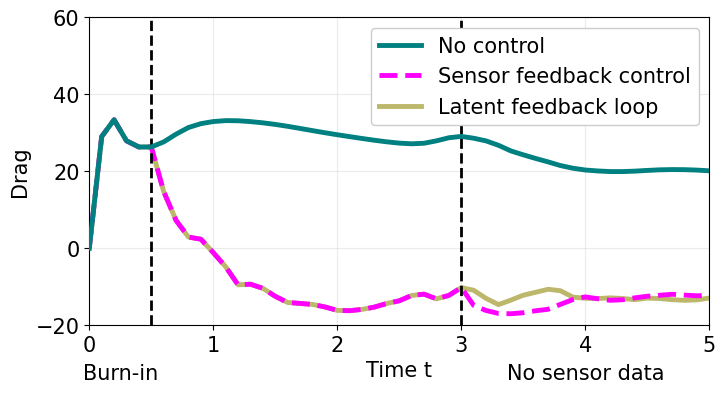}\put(-10,50){\bfseries (b)}\end{overpic}}
    \subfloat{
        \includegraphics[width=0.32\textwidth]{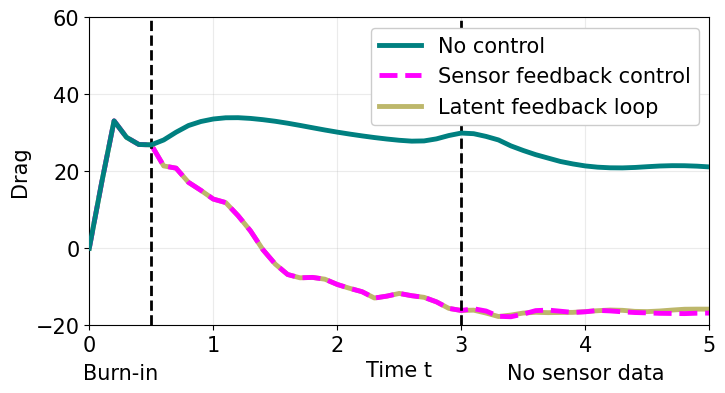}
      
    }
    \subfloat{
        \includegraphics[width=0.32\textwidth]{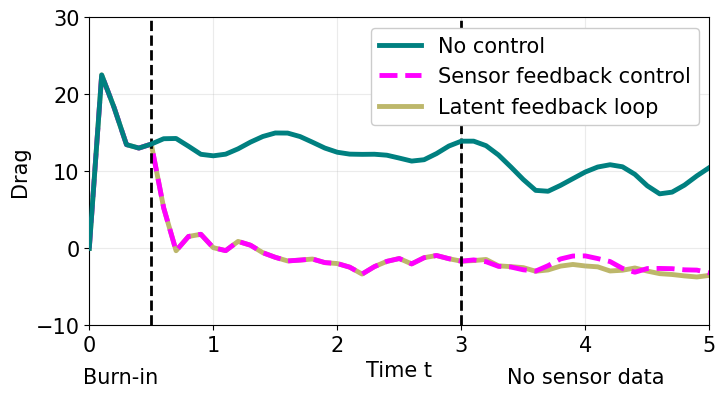}
      
    }
    \captionsetup{justification=raggedright, singlelinecheck=false}
    \caption{{\em Unsteady flow control}. {\bfseries (a)} Energy dissipated by the fluid in the uncontrolled and controlled settings with $\mu = -0.39, 0.39, -0.76$. {\bfseries (b)} Drag force on the obstacle in the uncontrolled and controlled settings with $\mu = -0.39, 0.39, -0.76$.}    
    \label{fig:flowshredc1}
\end{figure*}
\begin{equation}
    \begin{cases}
       \dot{\vec{v}} - \nu \Delta \vec{v} + (\vec{v} \cdot \nabla) \vec{v} + \nabla p = \vec{0}  \quad &\text{in} \ \Omega \times (0,T] \\
       \nabla \cdot \vec{v} = 0 \quad &\text{in} \ \Omega \times (0,T] \\
       \vec{v} = \vec{0} \quad &\text{on} \ \Gamma_{\text{obs}} \times (0,T]\\
       \vec{v} = \vec{u} \quad &\text{on} \ \Gamma_{\text{c}} \times (0,T]\\
       \vec{v} = \vec{v}_{\text{in}}(\boldsymbol{\mu})  \quad &\text{on} \ \Gamma_{\text{in}} \times (0,T]\\
       \vec{v} \cdot \vec{n} = 0  \quad &\text{on} \ \Gamma_{\text{walls}} \times (0,T] \\
       (\nu \nabla \vec{v} - p\mathbb{I})\vec{n} \cdot \vec{t} = 0  \quad &\text{on} \ \Gamma_{\text{walls}} \times (0,T] \\
       (\nu \nabla \vec{v} - p\mathbb{I})\vec{n} = \vec{0} \quad &\text{on} \ \Gamma_{\text{out}} \times (0,T]\\
       \vec{v} = \vec{0} &\text{in} \ \Omega \times \{t = 0\},
\end{cases}
\label{eq:unsteadyNS-OCPDLROM}
\end{equation}
where the kinematic viscosity is equal to $\nu=0.01$ and $\mathbb{I}$ is the identity matrix. Specifically, we consider homogeneous initial condition, homogeneous Dirichlet boundary condition on the rear of the obstacle $\Gamma_{\text{obs}}$, homogeneous Neumann boundary conditions on the outflow $\Gamma_{\text{out}}$, and free-slip boundary conditions on the walls $\Gamma_{\text{walls}}$. On the inflow boundary $\Gamma_{\text{in}}$, we take into account a stationary velocity datum
\begin{equation*}
\vec{v}_{\text{in}}(\mathbf{x}, \boldsymbol{\mu}) = [\gamma_{\text{in}}\cos(\alpha_{\text{in}}), 0.01 x_2 (20 - x_2) \gamma_\text{in} \sin(\alpha_{\text{in}})]^{\top}
\end{equation*}
where the angle of attack  $\alpha_{\text{in}}$ is regarded as scenario parameter, i.e., $\mu = \alpha_{\text{in}}$, while the inflow velocity intensity is set equal to $\gamma_{\text{in}}=10.0$, yielding a Reynolds number equal to $400$. 
\begin{figure*}
    \centering
    \subfloat{
    \begin{overpic}[width=0.47\textwidth]{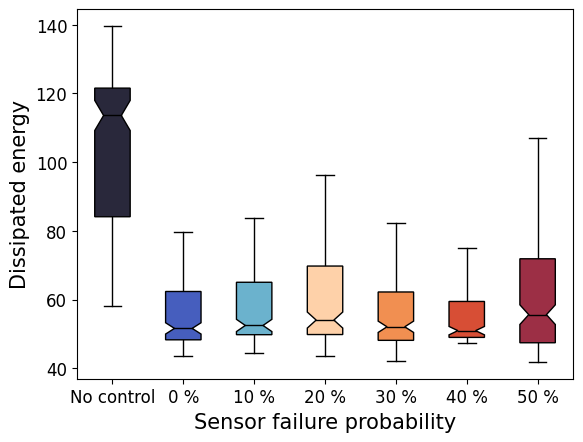}\put(-5,70){\bfseries (a)}\end{overpic}} \;
    \subfloat{
    \begin{overpic}[width=0.47\textwidth]{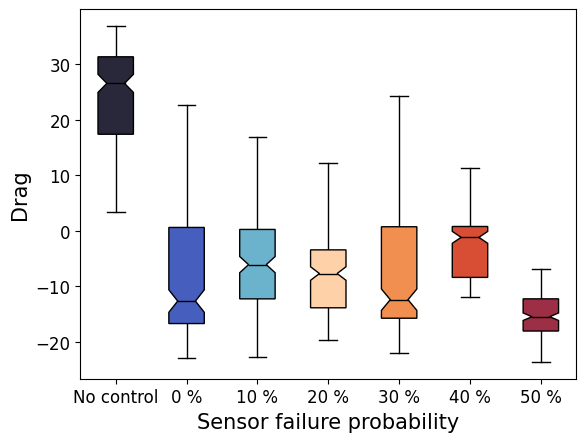}\put(0,70){\bfseries (b)}\end{overpic}}

    \captionsetup{justification=raggedright, singlelinecheck=false}
    \caption{{\em Unsteady flow control}. {\bfseries (a)} Energy dissipated by the fluid in the uncontrolled and controlled settings with latent feedback loop while considering $5$ different random parametric scenarios and different sensor failure probabilities. {\bfseries (b)} Drag force on the obstacle in the uncontrolled and controlled settings with latent feedback loop while considering $5$ different random parametric scenarios and different sensor failure probabilities.}    
    \label{fig:flowshredc2}
\end{figure*}

The control action $\vec{u}$ allows us to inject or absorb the fluid from the front portion of the cylinder boundary $\Gamma_{\text{c}}$. After discretizing Equation~\eqref{eq:unsteadyNS-OCPDLROM} with $\mathbb{P}_2-\mathbb{P}_1$ finite elements, the flow velocity and the boundary control action are described, respectively, by means of $N_v = 46874$ and $N_u=54$ degrees of freedom. After letting the system evolve for a burn-in period of $0.5$ seconds, the control is applied every $\Delta t = 0.1$ seconds until $T=5.0$ seconds, that is $N_t=45$. In this setting, we aim to minimize the energy dissipated by the fluid 
in multiple scenarios, that are different angles of attack $\mu = \alpha_{\text{in}} \in [-1.0,1.0]$. The dissipated energy can be computed as the $L^2$ norm of the flow velocity gradient, while regularizing terms avoid extremely expensive and energetic optimal control strategies~\cite{Tomasetto2024}.

After generating $50$ optimal solutions through \texttt{dolfin-adjoint}~\cite{Mitusch2019}, we double the snapshots thanks to horizontal symmetry, yielding $N_p = 100$ trajectories. In order to efficiently train the neural networks into play, we reduce the data dimensionality through POD with reduced dimensions equal to $200$ for the velocity and $20$ for the control. It is now possible to train the SHRED-ROM policy to predict distributed control actions starting from the temporal history of $L=20$ values of $N_s=3$ sensors monitoring the horizontal velocity at as many locations in the channel. Note that the sensor placement is chosen randomly, and comparable results can be retrieved exploiting different sensor locations, taking into account more sensors, as well as measuring different related quantities such as, e.g., the pressure~\cite{tomasetto2025shredrom}. After training, SHRED-ROM is capable of accurately reconstructing boundary controls from state sensor values, with a mean relative error in test scenarios equal to $2.68\%$. 

Whenever optimal state data are available, one can augment the SHRED-ROM output in order to predict both distributed optimal controls and the corresponding high-dimensional optimal state. Specifically, we train the sequence and decoder model of SHRED-ROM to return the $220$ POD coefficients of state and control, while we recover the corresponding full-order variables through POD decoding, that is
\[
\begin{bmatrix}
    {\mathbf{u}}_k^{\boldsymbol{\mu}}
    \\
    {\mathbf{v}}_k^{\boldsymbol{\mu}}
\end{bmatrix}
\approx
\begin{bmatrix}
    \hat{\mathbf{u}}_k^{\boldsymbol{\mu}}
    \\
    \hat{\mathbf{v}}_k^{\boldsymbol{\mu}}
\end{bmatrix}
=
\begin{bmatrix}
    \boldsymbol{\Psi}_u & \boldsymbol{0}
    \\
    \boldsymbol{0} & \boldsymbol{\Psi}_v
\end{bmatrix}
\mathbf{f}_X(\mathbf{f}_T(\mathbf{s}_{k-L:k}^{\boldsymbol{\mu}}))
\]
for $k=1,\ldots,N_t$, where $\boldsymbol{\Psi}_u$ and $\boldsymbol{\Psi}_v$ denote the POD modes for control and velocity, respectively. Therefore, beyond predicting the optimal control action to apply to the dynamical system, it is also possible to approximate the underlying high-dimensional fluid velocity, with a mean relative error on test data equal to $2.38 \%$. Figure~\ref{fig:flowshredc} shows the uncontrolled velocity, the target optimal pair computed through a high-fidelity OCP solver, the flow velocity under the SHRED-ROM policy, and the SHRED-ROM optimal pair prediction. The proposed sensor-based feedback controller is able to effectively suppress the vortex shedding in the cylinder wake and to predict the evolution of the high-dimensional controlled dynamics, resulting in a minimization of the dissipated energy and the drag force on the cylinder, as further assessed by the results in Figure~\ref{fig:flowshredc1}.

The lack of sensor values in the online phase prevents real-time adaptations of the control strategy, thereby compromising efficiency and robustness, while possibly leading to unsafe operating conditions. To overcome this limitation, we propose a loop closure at the latent level through a latent sensor forecaster $\varphi_z$ capable of replacing the missing sensor measurements with accurate approximations. After training $\varphi_z$ with lag $L=20$, the latent sensor forecaster is able to accurately forecast the sensor evolution over time starting from past sensor measurements and latent control variables, with a mean relative error on test data equal to $0.19\%$. Figure~\ref{fig:flowshredc} shows the controlled velocity and the corresponding optimal pair prediction in three test scenarios when switching off the sensors from $t=3$ onward, thus employing the latent sensor forecaster to predict the future sensor values in autoregressive mode. Moreover, Figure~\ref{fig:flowshredc1} displays the resulting flow energy dissipation and drag, with minimal discrepancy with respect to the controlled setting relying on sensor data at every time step. Figure~\ref{fig:flowshredc2} presents, instead, an ablation study on the dissipated energy and the drag force for different sensor failure probabilities. In particular, we exploit the sensor-based feedback controller with latent feedback loop to minimize the dissipated energy in $5$ different scenarios randomly sampled in the parameter space $\mathcal{P}$. At every time step, each sensor may experience a malfunction with a certain probability, thus requiring the latent sensor forecaster to predict its evolution. Despite remarkably high failure probabilities up to $50 \%$, the proposed closed-loop control strategy is able to significantly decrease the dissipated energy with respect to the uncontrolled setting, with minimal differences compared to the case monitoring sensor data at every time step. 


\clearpage
\subsection*{Double gyre flow tracking}

\begin{figure*}
    \centering
    \begin{sideways}
    \makebox[0pt][l]{\hspace{-3.12cm}
    \begin{minipage}{4cm}
    \centering
    {\bfseries \small TARGET} \\   {\bfseries \small VELOCITY} \end{minipage}}
    \end{sideways}  \subfloat[\shortstack{$1^{st}$ test scenario \\ 0.5 seconds}]{
        \includegraphics[width=0.22\textwidth]{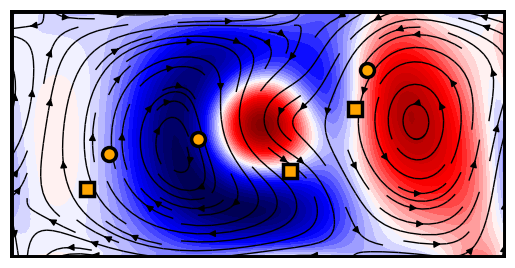}
      
    }
    \subfloat[\shortstack{$1^{st}$ test scenario \\ 2.5 seconds}]{
        \includegraphics[width=0.22\textwidth]{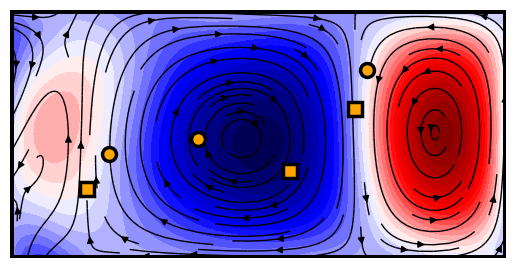}
      
    }
    \subfloat[\shortstack{$2^{nd}$ test scenario \\ 0.5 seconds}]{
        \includegraphics[width=0.22\textwidth]{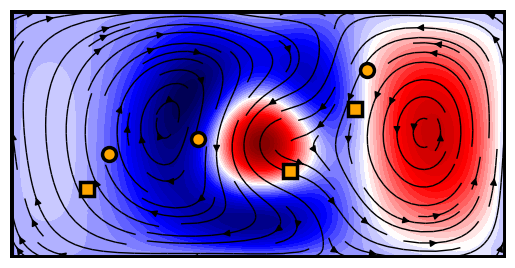}
      
    }
    \subfloat[\shortstack{$2^{nd}$ test scenario \\ 2.5 seconds}]{
        \includegraphics[width=0.22\textwidth]{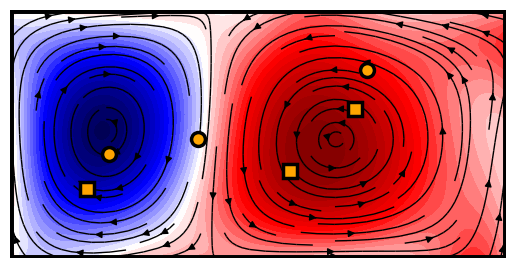}
      
    }

    \begin{sideways}
    \makebox[0pt][l]{\hspace{-3.12cm}
    \begin{minipage}{4cm}
    \centering
    {\bfseries \small SHRED-ROM} \\   {\bfseries \small VELOCITY} \end{minipage}}
    \end{sideways}  \subfloat{
        \includegraphics[width=0.22\textwidth]{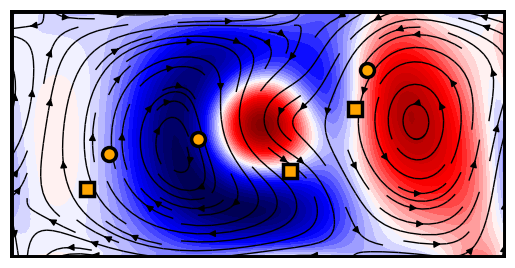}
      
    }
    \subfloat{
        \includegraphics[width=0.22\textwidth]{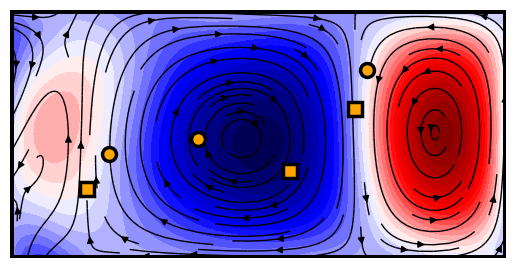}
      
    }
    \subfloat{
        \includegraphics[width=0.22\textwidth]{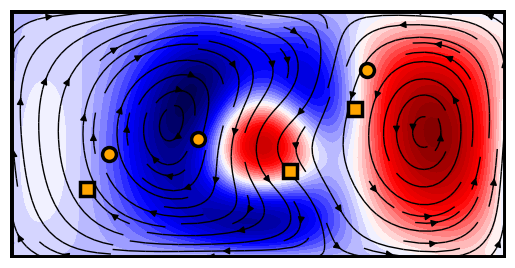}
      
    }
    \subfloat{
        \includegraphics[width=0.22\textwidth]{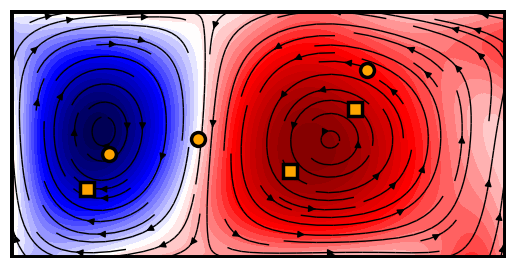}
      
    }

    \vspace{-0.25cm}
    \hspace{0.6cm}
    \subfloat{
        \includegraphics[width=0.22\textwidth]{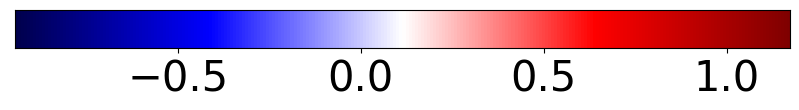}
      
    }
    \subfloat{
        \includegraphics[width=0.22\textwidth]{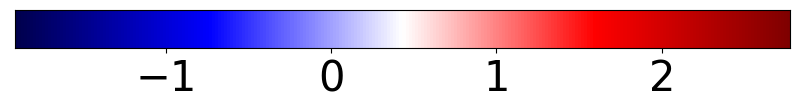}
      
    }
        \subfloat{
        \includegraphics[width=0.22\textwidth]{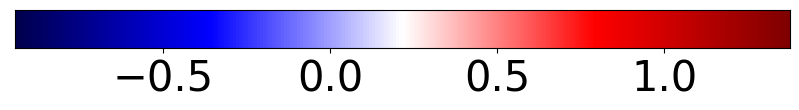}
      
    }
        \subfloat{
        \includegraphics[width=0.22\textwidth]{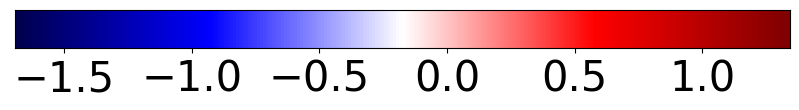}
      
    }

\vspace{-0.4cm}
    \begin{sideways}
    \makebox[0pt][l]{\hspace{-3.12cm}
    \begin{minipage}{4cm}
    \centering
    {\bfseries \small TARGET} \\   {\bfseries \small CONTROL} \end{minipage}}
    \end{sideways}  \subfloat{
        \includegraphics[width=0.22\textwidth]{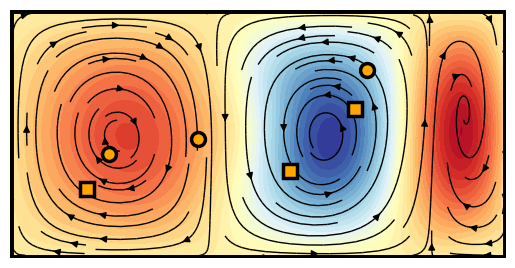}
      
    }
    \subfloat{
        \includegraphics[width=0.22\textwidth]{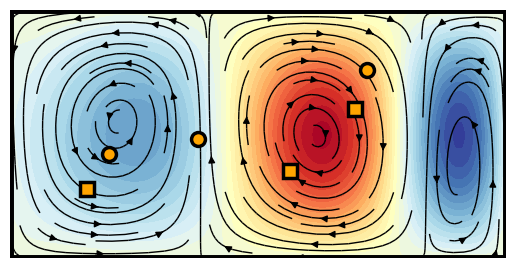}
      
    }
    \subfloat{
        \includegraphics[width=0.22\textwidth]{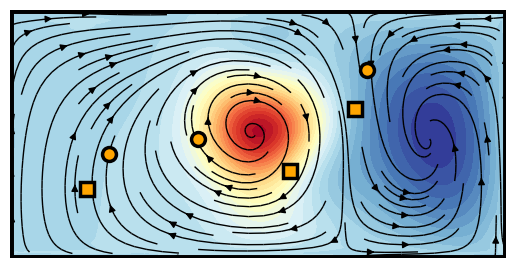}
      
    }
    \subfloat{
        \includegraphics[width=0.22\textwidth]{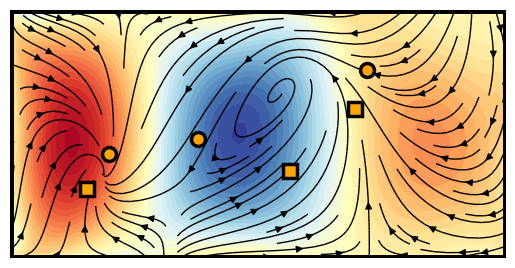}
      
    }

    \begin{sideways}
    \makebox[0pt][l]{\hspace{-3.12cm}
    \begin{minipage}{4cm}
    \centering
    {\bfseries \small SHRED-ROM} \\   {\bfseries \small CONTROL} \end{minipage}}
    \end{sideways}  \subfloat{
        \includegraphics[width=0.22\textwidth]{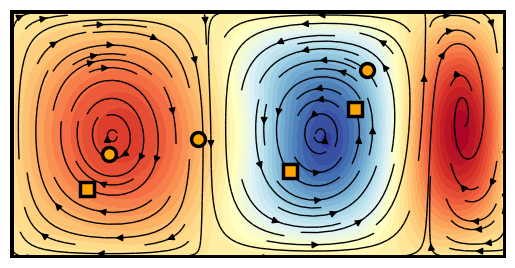}
      
    }
    \subfloat{
        \includegraphics[width=0.22\textwidth]{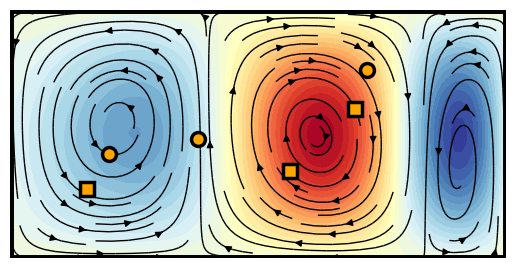}
      
    }
    \subfloat{
        \includegraphics[width=0.22\textwidth]{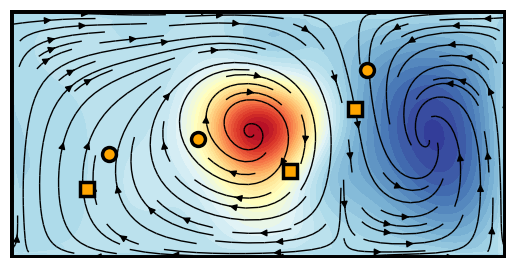}
      
    }
    \subfloat{
        \includegraphics[width=0.22\textwidth]{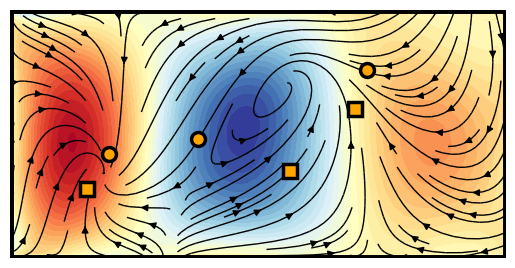}
      
    }   

\vspace{-0.25cm}
    \hspace{0.6cm}
    \subfloat{
        \includegraphics[width=0.22\textwidth]{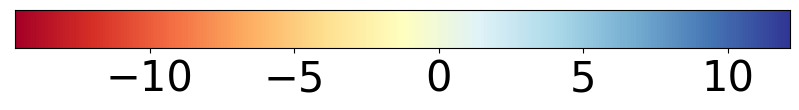}
      
    }
    \subfloat{
        \includegraphics[width=0.22\textwidth]{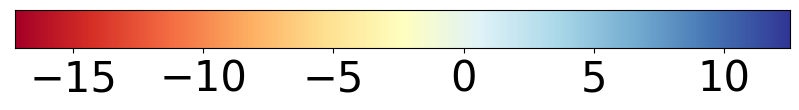}
      
    }
        \subfloat{
        \includegraphics[width=0.22\textwidth]{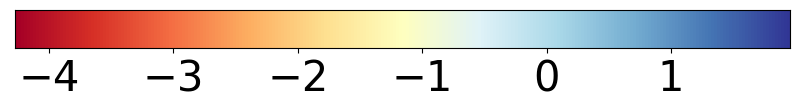}
      
    }
        \subfloat{
        \includegraphics[width=0.22\textwidth]{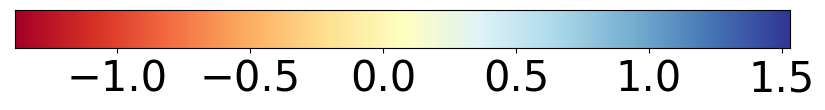}
      
    }

\captionsetup{justification=raggedright, singlelinecheck=false}
    \caption{{\em Double gyre flow tracking}. Optimal target velocity (first row), SHRED-ROM optimal velocity prediction (second row), optimal target control (third row) and SHRED-ROM optimal control prediction (fourth row) in two different test scenarios related to $\boldsymbol{\mu} = [0.49, 5.4]^\top$ and $\boldsymbol{\mu} = [0.22, 1.47]^\top$ at $t=0.5$ and $t=2.5$ seconds. The fixed sensors monitoring pressure and reference velocity are depicted with, respectively, orange dots and squares. The velocity and control on $\Omega$ are depicted through a scalar field with colors corresponding to its vorticity, with the streamlines in black.}
    \label{fig:2gyre}
\end{figure*}

\begin{figure*}
    \centering
    \subfloat[\hspace{0.4cm} \shortstack{$1^{st}$ test scenario \\ Pressure sensor data}]{
        \includegraphics[width=0.25\textwidth]{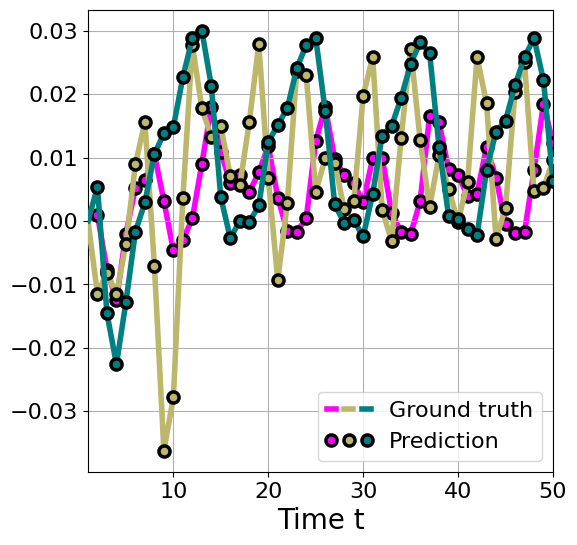}
      
    }
    \subfloat[\hspace{0.4cm} \shortstack{$1^{st}$ test scenario \\ Reference sensor data}]{
        \includegraphics[width=0.25\textwidth]{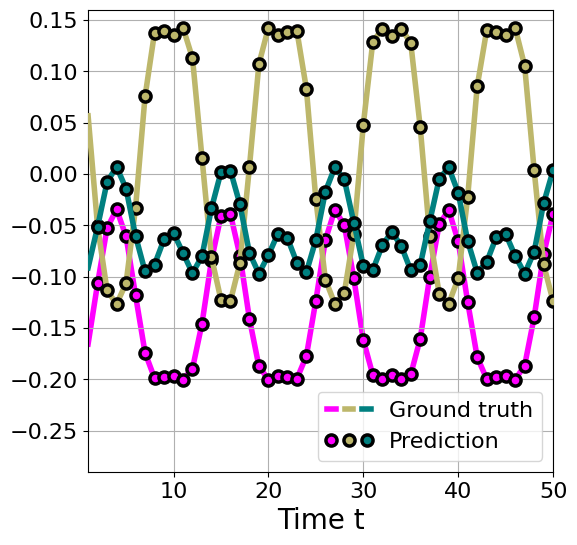}
      
    }
        \subfloat[\hspace{0.4cm} \shortstack{$2^{nd}$ test scenario \\ Pressure sensor data}]{
        \includegraphics[width=0.25\textwidth]{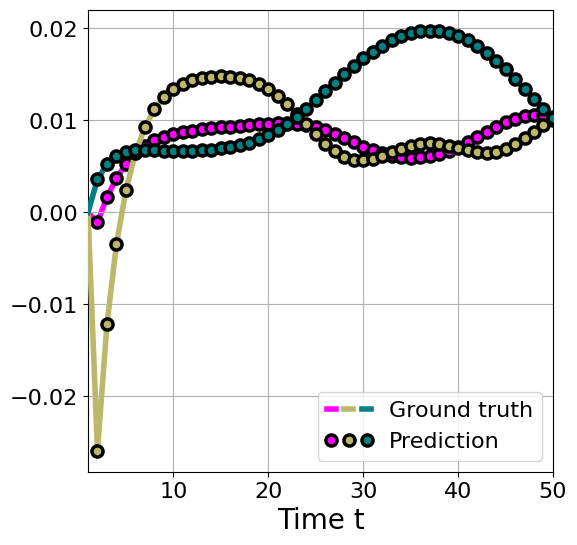}
      
    }
        \subfloat[\hspace{0.4cm} \shortstack{$2^{nd}$ test scenario \\ Reference sensor data}]{
        \includegraphics[width=0.25\textwidth]{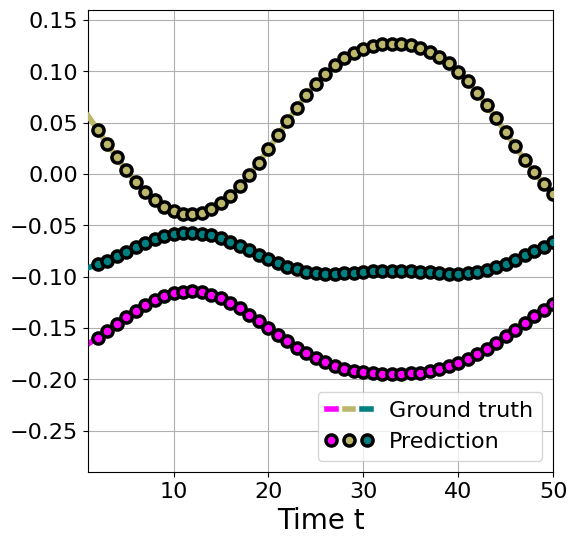}
      
    }
    \captionsetup{justification=raggedright, singlelinecheck=false}
    \caption{{\em Double gyre flow tracking}. Ground truth pressure sensor data (first and third panel) and reference horizontal velocity sensor data (second and fourth panel) with latent sensor forecaster predictions in two different test scenarios related to $\boldsymbol{\mu} = [0.49, 5.4]^\top$ (first and second panel) and $\boldsymbol{\mu} = [0.22, 1.47]^\top$ (third and fourth panel).}
    \label{fig:2gyre_lfs}
\end{figure*}
In this section, we exploit SHRED-ROM to solve a high-dimensional and parametric tracking problem. In particular, we consider an incompressible fluid in the rectangular domain $\Omega = (0,2) \times (0,1)$, whose dynamics is governed by the unsteady NS equations
\begin{equation}
    \begin{cases}
       \dot{\vec{v}} - \nu \Delta \vec{v} + (\vec{v} \cdot \nabla) \vec{v} + \nabla p = \vec{u}  \quad &\text{in} \ \Omega \times (0,T] \\
       \nabla \cdot \vec{v} = 0 \quad &\text{in} \ \Omega \times (0,T] \\
       (\nu \nabla \vec{v} - p\mathbb{I})\vec{n} = \vec{0} \quad &\text{on} \ \partial \Omega \times (0,T]\\
       \vec{v} = \vec{v}_0 &\text{in} \ \Omega \times \{t = 0\},
\end{cases}
\label{eq:unsteadyNS}
\end{equation}
with $\nu = 0.01$ and homogeneous Neumann boundary conditions. Starting from the initial velocity configuration
\[
\vec{v}_0(\mathbf{x}) = \left[\dfrac{\partial \psi}{\partial x_2},-\dfrac{\partial \psi}{\partial x_1}\right]^\top,
\]
where the stream function $\psi(\mathbf{x}) = \psi(x_1, x_2) = 0.1 e^{-16((x_1 - 1.0)^2 + (x_2 - 0.5)^2)}$ entails a vortex centered in the domain, the goal is to steer the fluid dynamics in order to track a reference velocity $\bar{v}$. In other words, we aim to minimize the $L^2$ distance between the NS velocity $\vec{v}$ and the target one $\bar{v}$. To do so, we employ a distributed, time-dependent and vector-valued source term $\vec{u}$ as control action. In the following, the target is represented by the double gyre flow~\cite{SHADDEN2005271}, a time-dependent model for the interaction of two counter-rotating vortices that can lead to chaotic particle trajectories. Specifically, the double gyre flow velocity is defined as
\begin{equation*}
\bar{v}(\mathbf{x}, \boldsymbol{\mu}) = 
\begin{bmatrix}
- \pi I \sin\left( \pi f \right) \cos\left( \pi x_2 \right)
\\
\pi I \cos\left( \pi f \right) \sin\left( \pi x_2 \right) \frac{\partial f}{\partial x_1}
\end{bmatrix},
\end{equation*}
where $I=0.1$ is the intensity coefficient, $f = f(x_1,  t) = \epsilon \sin(\omega t) x_1^2 + (1 - 2\epsilon \sin(\omega t)) x_1$, while $\epsilon$ and $\omega$ are the perturbation amplitude and the frequency of the oscillation, respectively, which are regarded as scenario parameters $\boldsymbol{\mu} = [\epsilon, \omega]^\top$.

After discretizing the unsteady NS equations with $\mathbb{P}_2-\mathbb{P}_1$ finite elements, we end up with $5151$ degrees of freedom for the pressure, while the vector-valued fields have dimension equal to $N_v = N_u = 40602$. Regarding the time discretization, we consider a uniform grid within the horizon $[0,T]$, with final time $T=5$ seconds and time step $\Delta t=0.1$ seconds, that is $N_t=50$.

Instead of relying on optimal control solvers to generate training snapshots, we here take into account the feedback control law proposed by Strazzullo et al.~\cite{strazzullo2025newfeedbackcontroladaptive}, which extends the results by Gunzburger and Manservisi~\cite{gunzburger1991analysis}. In particular, the expert behaviour is given by the source term
\[
\vec{u} = \dot{\bar{v}} - \nu \Delta \bar{v} + (\bar{v} \cdot \nabla) \bar{v}  + (\vec{v} - \bar{v}) \nabla \bar{v} -  (\vec{v} - \bar{v}),
\]
which enables velocity tracking with exponential convergence over time. Note that the control optimality is guaranteed even after space-time discretization~\cite{strazzullo2025newfeedbackcontroladaptive}. Through the control formula above, we generate $N_p=100$ optimal control actions, along with the corresponding velocity and pressure fields, for as many scenario parameters randomly sampled within $\mathcal{P} = [0.0, 0.5] \times [0.25\pi, 2\pi]$. The velocity and control snapshots are then reduced through POD in order to enable SHRED-ROM compressive training. In particular, we consider $15$ modes for each variable component, for a total of $60$ POD modes.

SHRED-ROM is now utilized to predict in real-time the distributed and parametric control actions and, similarly to the previous test case, the corresponding velocity fields. The input is represented by the temporal history of $6$ sensors randomly placed in the domain, half of which monitor the pressure and half of which measure the horizontal velocity of the reference flow. Note that, while the former set of sensors is crucial to inform the controller about the system configuration, the latter is needed to specify which instance of double gyre flow to track, without requiring complete knowledge of the scenario parameters. The lag parameter is set equal to $L=25$, which covers two periods of the double gyre flow at the minimum frequency considered. After training SHRED-ROM to predict the $60$ POD coefficients of the variables of interest, we are able to accurately reconstruct test snapshots with a mean relative error equal to $7.47 \%$ for the control and $3.00 \%$ for the velocity. Figure~\ref{fig:2gyre} qualitatively assesses the performance of SHRED-ROM when predicting control and velocity snapshots in two test cases, relying solely on the limited sensor readings available. To cope with missing sensor values online, we also train a latent sensor forecaster to predict sensor values starting from their own $L=25$ past values and the corresponding control latent variables, resulting in a mean relative error on test data equal to $0.69 \%$. Figure~\ref{fig:2gyre_lfs} displays the goodness of $\varphi_z$ in forecasting ground truth sensor data in the test set.

\begin{figure*}
    \centering
    \begin{sideways}
    \makebox[0pt][l]{\hspace{-3.12cm}
    \begin{minipage}{4cm}
    \centering
    {\bfseries \small NO} \\   {\bfseries \small CONTROL} \end{minipage}}
    \end{sideways}  \subfloat[0.5 seconds]{
        \includegraphics[width=0.22\textwidth]{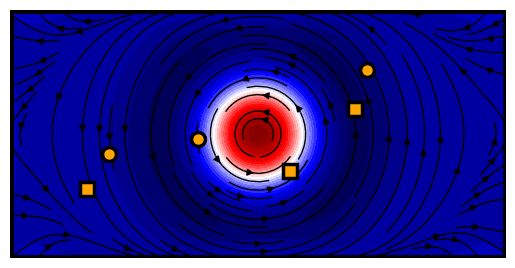}
      
    }
    \subfloat[2.0 seconds]{
        \includegraphics[width=0.22\textwidth]{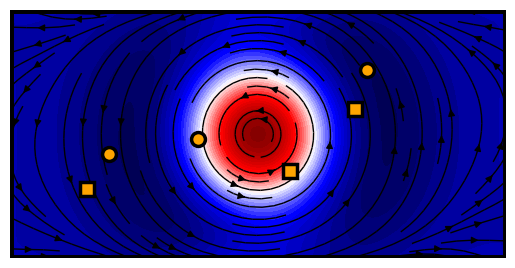}
      
    }
    \subfloat[3.5 seconds]{
        \includegraphics[width=0.22\textwidth]{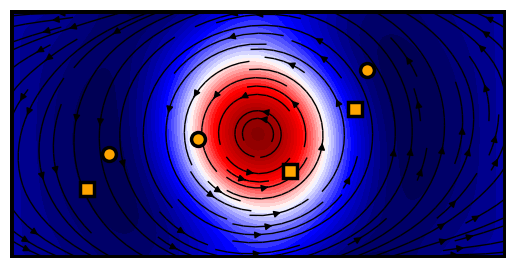}
      
    }
    \subfloat[5.0 seconds]{
        \includegraphics[width=0.22\textwidth]{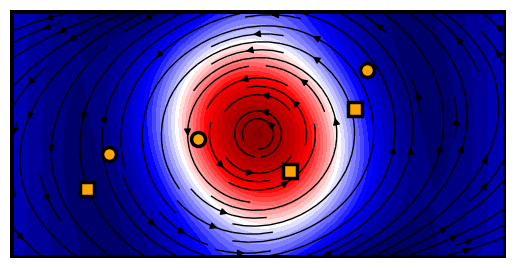}
      
    }

        \begin{sideways}
    \makebox[0pt][l]{\hspace{-3.12cm}
    \begin{minipage}{4cm}
    \centering
    {\bfseries \small REFERENCE} \\   {\bfseries \small VELOCITY} \end{minipage}}
    \end{sideways}  \subfloat{
        \includegraphics[width=0.22\textwidth]{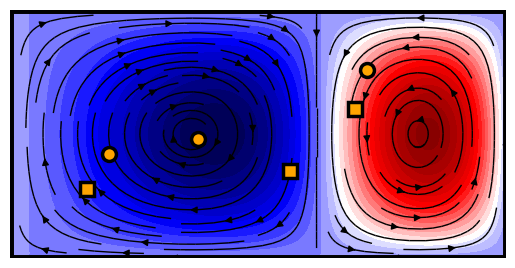}
      
    }
    \subfloat{
        \includegraphics[width=0.22\textwidth]{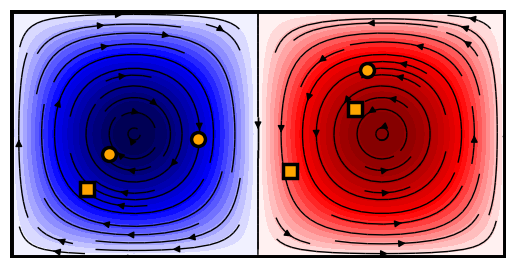}
      
    }
    \subfloat{
        \includegraphics[width=0.22\textwidth]{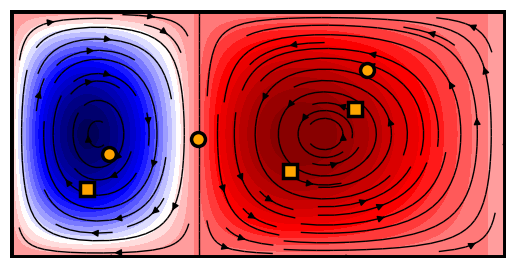}
      
    }
    \subfloat{
        \includegraphics[width=0.22\textwidth]{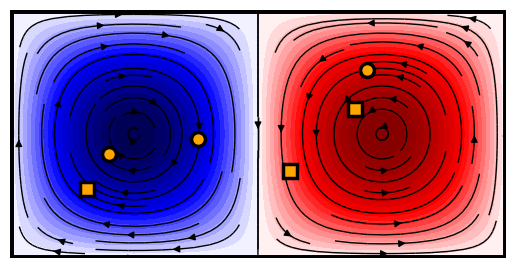}
      
    }

 \begin{sideways}
    \makebox[0pt][l]{\hspace{-4.4cm}
    \begin{minipage}{4.2cm}
    \centering
    {\bfseries \small CONTROLLED} \\  {\bfseries \small VELOCITY} \end{minipage}}
    \end{sideways}\subfloat{
        \includegraphics[width=0.22\textwidth]{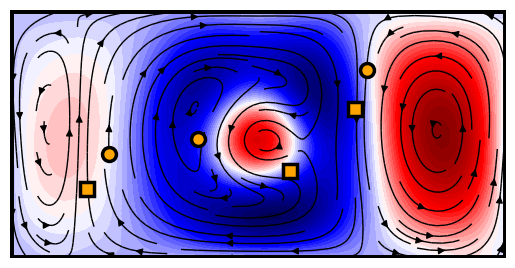}
      
    }
    \subfloat{
        \includegraphics[width=0.22\textwidth]{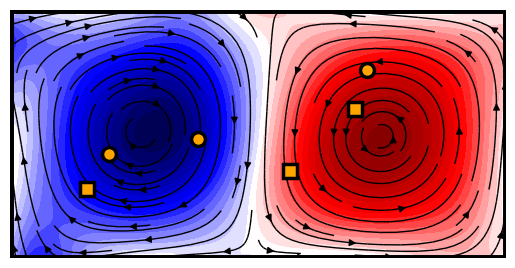}
      
    }
    \subfloat{
        \includegraphics[width=0.22\textwidth]{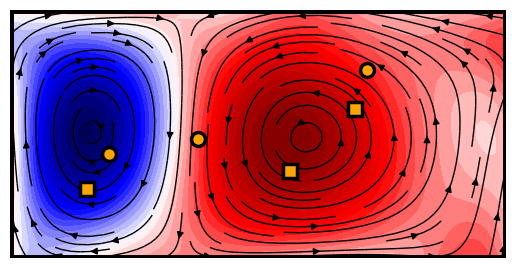}
      
    }
    \subfloat{
        \includegraphics[width=0.22\textwidth]{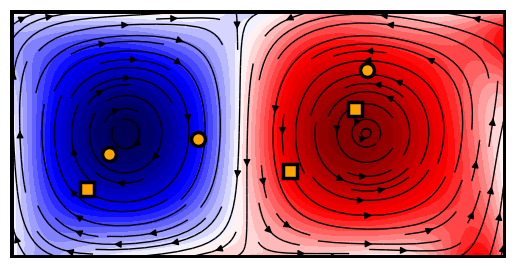}
      
    }

    \begin{sideways}
    \makebox[0pt][l]{\hspace{-3cm}
    \begin{minipage}{4cm}
    \centering
    \vphantom{\bfseries \small SHRED-ROM} \\   \vphantom{\bfseries \small PREDICTION} \end{minipage}}
    \end{sideways} 
    \subfloat{
        \includegraphics[width=0.22\textwidth]{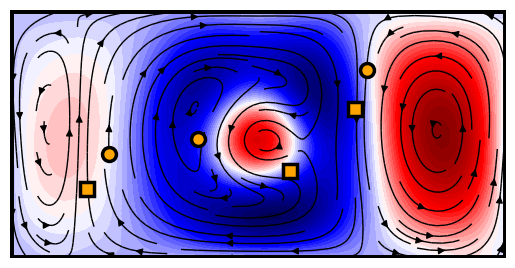}
      
    }
    \subfloat{
        \includegraphics[width=0.22\textwidth]{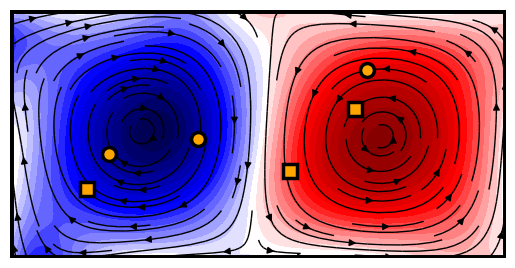}
      
    }
    \subfloat{
        \includegraphics[width=0.22\textwidth]{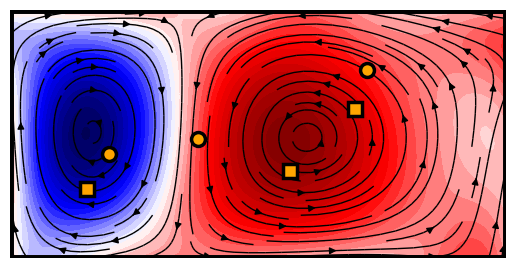}
      
    }
    \subfloat{
        \includegraphics[width=0.22\textwidth]{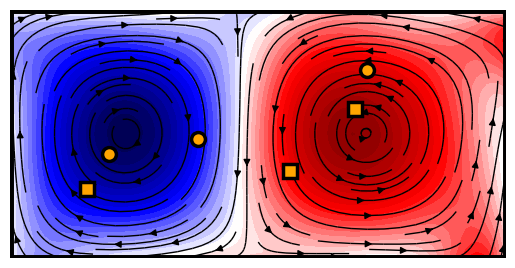}
      
    }

    \vspace{-0.25cm}
    \hspace{0.6cm}
    \subfloat{
        \includegraphics[width=0.22\textwidth]{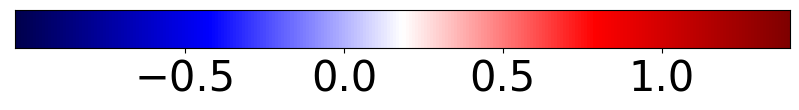}
      
    }
    \subfloat{
        \includegraphics[width=0.22\textwidth]{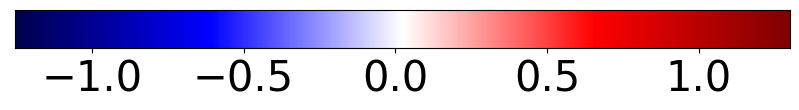}
      
    }
        \subfloat{
        \includegraphics[width=0.22\textwidth]{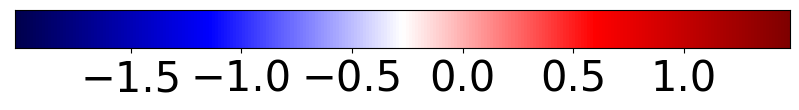}
      
    }
        \subfloat{
        \includegraphics[width=0.22\textwidth]{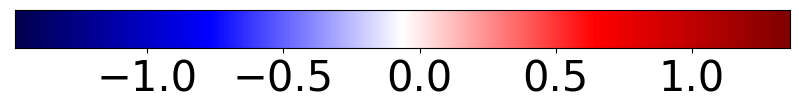}
      
    }    

    \begin{sideways}
    \makebox[0pt][l]{\hspace{-4.5cm}
    \begin{minipage}{4.2cm}
    \centering
    {\bfseries \small SHRED-ROM} \\  {\bfseries \small CONTROL} \end{minipage}}
    \end{sideways}
    \subfloat{
        \includegraphics[width=0.22\textwidth]{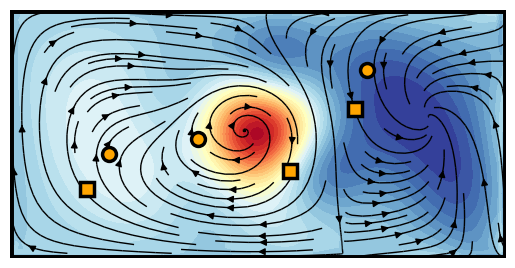}
      
    }
    \subfloat{
        \includegraphics[width=0.22\textwidth]{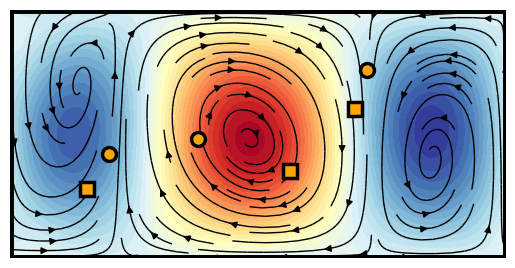}
      
    }
    \subfloat{
        \includegraphics[width=0.22\textwidth]{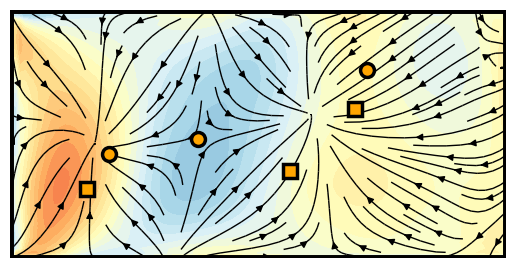}
      
    }
    \subfloat{
        \includegraphics[width=0.22\textwidth]{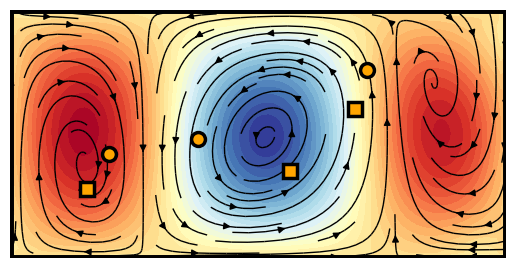}
      
    }

    \begin{sideways}
    \makebox[0pt][l]{\hspace{-3cm}
    \begin{minipage}{4cm}
    \centering
    \vphantom{\bfseries \small SHRED-ROM} \\   \vphantom{\bfseries \small PREDICTION} \end{minipage}}
    \end{sideways} 
    \subfloat{
        \includegraphics[width=0.22\textwidth]{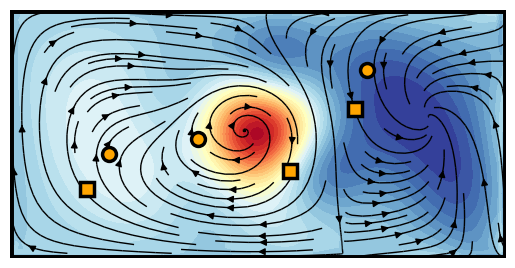}
      
    }
    \subfloat{
        \includegraphics[width=0.22\textwidth]{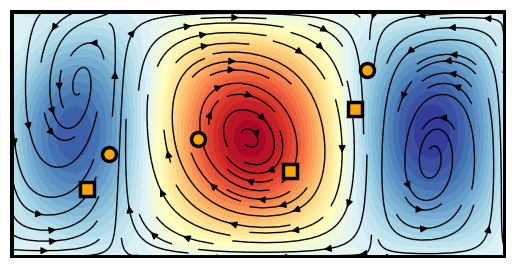}
      
    }
    \subfloat{
        \includegraphics[width=0.22\textwidth]{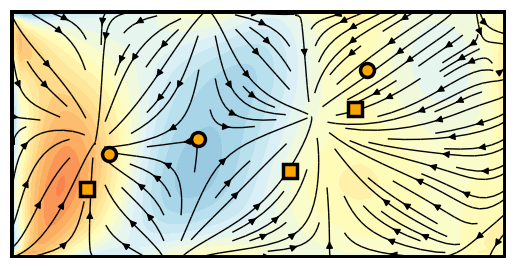}
      
    }
    \subfloat{
        \includegraphics[width=0.22\textwidth]{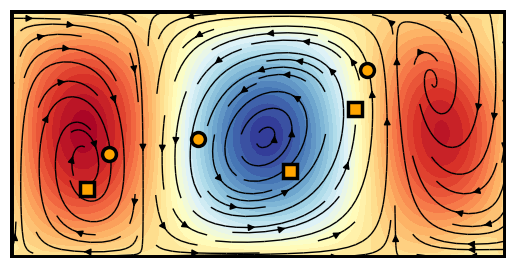}
      
    }
    
    \vspace{-0.25cm}
    \hspace{0.6cm}
    \subfloat{
        \includegraphics[width=0.22\textwidth]{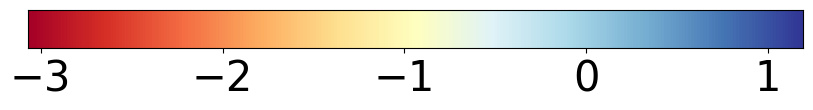}
      
    }
    \subfloat{
        \includegraphics[width=0.22\textwidth]{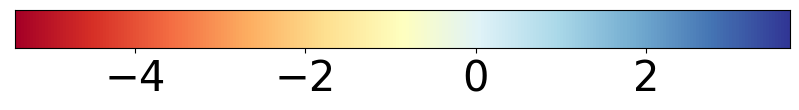}
      
    }
        \subfloat{
        \includegraphics[width=0.22\textwidth]{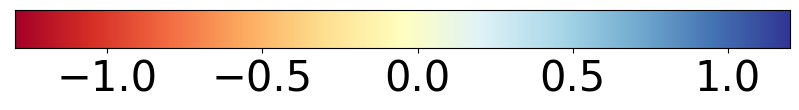}
      
    }
        \subfloat{
        \includegraphics[width=0.22\textwidth]{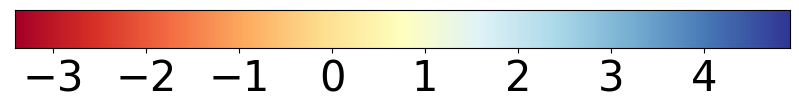}
      
    } 
\captionsetup{justification=raggedright, singlelinecheck=false}
    \caption{{\em Double gyre flow tracking}. Uncontrolled velocity (first row), reference velocity (second row), controlled velocity and optimal control prediction with sensor-based feedback loop (third and fifth row), controlled velocity and optimal control prediction with latent feedback loop (fourth and sixth row) in the test scenario related to $\boldsymbol{\mu} = [0.25, \pi]^\top$ at $t=0.5, 2.0, 3.5, 5.0$ seconds. The fixed sensors monitoring pressure and reference velocity are depicted with, respectively, orange dots and squares. The velocity and control on $\Omega$ are depicted through a scalar field with colors corresponding to its vorticity, with the streamlines in black.}
    \label{fig:2gyre_test}
\end{figure*}

\begin{figure*}
    \centering
     \subfloat{
        \includegraphics[width=0.33\textwidth]{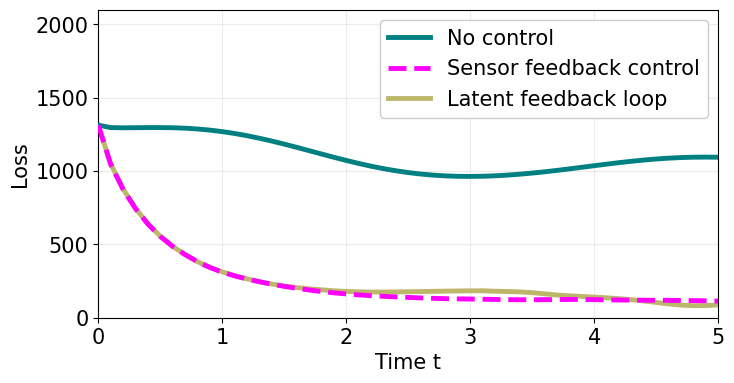}
      
    }
    \subfloat{
        \includegraphics[width=0.33\textwidth]{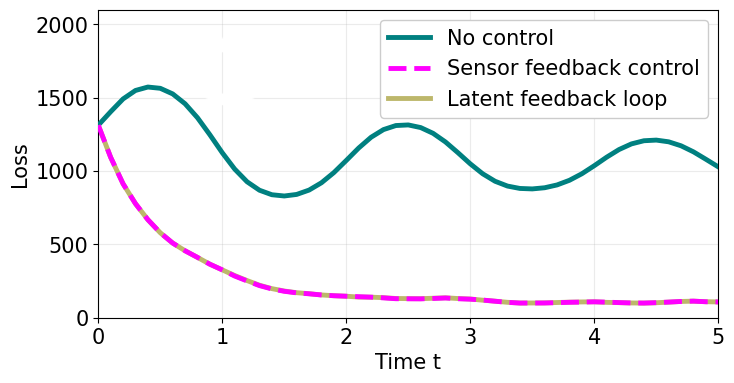}
      
    }    
    \subfloat{
        \includegraphics[width=0.33\textwidth]{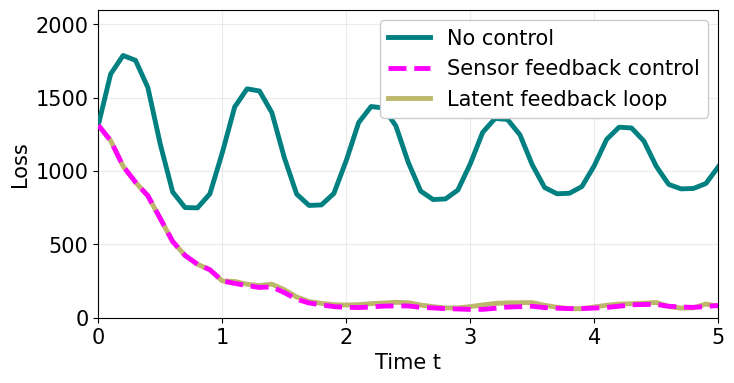}
      
    }    
\captionsetup{justification=raggedright, singlelinecheck=false}
    \caption{{\em Double gyre flow tracking}. Loss function in the uncontrolled and controlled settings with $\boldsymbol{\mu} = [0.1, 0.5\pi]^\top$ (first panel), $\boldsymbol{\mu} = [0.25, \pi]^\top$ (second panel) and $\boldsymbol{\mu} = [0.4, 2 \pi]^\top$ (third panel).}
    \label{fig:2gyre_loss}
\end{figure*}

The SHRED-ROM policy can be now employed to control test scenarios. In addition to the setting where sensor data are always available in time, we simulate a signal transmission problem every $1$ second, thus taking advantage of the latent sensor forecaster to close the loop at the latent level through the prediction of pressure and reference horizontal velocity sensor values. Figure~\ref{fig:2gyre_test} shows the uncontrolled dynamics, the double gyre reference flow, the velocity under SHRED-ROM policy and the corresponding SHRED-ROM control, both with and without sensor failures, related to $\boldsymbol{\mu} = [0.25, \pi]^\top$. Moreover, Figure~\ref{fig:2gyre_loss} displays the loss functions over time in the uncontrolled and controlled settings in three different test scenario parameters. SHRED-ROM is thus capable of rapidly steering the fluid dynamics from the initial single vortex to the double gyre reference flow, despite the distributed control actions required over time and the new scenarios never seen during training. 

\section{CONCLUSIONS}
In this work, we propose a parametric feedback control strategy in the low-data limit, agnostic to parameter values. Specifically, in the context of imitation learning, we exploit SHRED-ROM to mimic expert demonstrations and predict distributed control actions in multiple scenarios, relying solely on limited state sensor readings. Moreover, it is readily possible to enlarge the model output with the corresponding controlled state dynamics, as well as other coupled fields of interest. SHRED-ROM is particularly appealing for control tasks as {\em (i)} it requires very limited and sparse sensor measurements, {\em (ii)} it is independent on sensor placement and agnostic to parameter values, and {\em (iii)} it can be efficiently trained with laptop-level computing thanks to dimensionality reduction techniques such as POD. As demonstrated in Section~\ref{sec:test}, after SHRED-ROM training, it is possible to control challenging high-dimensional and parametric systems, designing effective optimal control strategies in real-time.

To account for sensor malfunctions and signal transmission problems, we propose a latent sensor forecaster to close the loop at the latent level. In particular, starting from past sensor values and control actions at the latent level, we employ a LSTM to predict the next sensor measurements, thus enabling continuous monitoring and control of the dynamical system at hand. As demonstrated throughout the test cases in Section~\ref{sec:test}, the latent feedback loop allows us to control the system even when coping with very high probability sensor failures or frequent signal transmission failures.  

The proposed sensor-based feedback control strategy may be extended in multiple directions in future works. For instance, uncertainty quantification and robust predictions~\cite{gao2026uqshreduncertaintyquantificationshallow, riva2024robuststateestimationpartial} may be helpful to account for noisy sensors, as well as to understand when to trust the controller and when fine-tuning is needed. Moreover, latent dynamics modeling~\cite{gao2025sparse, yermakov2025tshredsymbolicregressionregularization}, especially with a view of parsimonious and interpretable models~\cite{kutz2022parsimony}, would be helpful for control predictions over longer time horizons in extrapolation regimes. Finally, reinforcement learning routines could be helpful for policy fine-tuning, thus further enhancing control robustness under external disturbances, or for directly training the SHRED-ROM policy, without requiring optimal examples from expert demonstrators.
\\ 

\section*{DATA AND CODE AVAILABILITY}

The data necessary to replicate the test cases detailed in Section~\ref{sec:test} are publicly available at \url{https://doi.org/10.5281/zenodo.20627878}

The implementation of the proposed methodology and the code necessary to replicate the test cases detailed Section~\ref{sec:test} are publicly available at \url{https://github.com/MatteoTomasetto/shred-c}
\\

\section*{ACKNOWLEDGEMENTS} AM acknowledges the Project “Reduced Order Modeling and Deep Learning for the real-time approximation of PDEs (DREAM)” (Starting Grant No. FIS00003154), funded by the Italian Science Fund (FIS) - Ministero dell'Università e della Ricerca and the project FAIR (Future Artificial Intelligence Research), funded by the NextGenerationEU program within the PNRR-PE-AI scheme (M4C2, Investment 1.3, Line on Artificial Intelligence).

\vfill
\onecolumngrid
\vspace{5mm}
\noindent\centering\rule{0.5\textwidth}{1pt}
\vspace{5mm}
\twocolumngrid
\renewcommand{\bibsection}{}
\bibliographystyle{abbrv}
\bibliography{biblio,biblio1,biblio2,biblio3,biblio4,biblio5,biblio6}

\end{document}